# Multi-Agent Retrieval-Augmented Causality Assessment (MARCA) of Individual Case Safety Reports in Pharmacovigilance

**Daisy Rae Lopez [1†], Angela Boccia [1,2†], Maurizio Sessa [1]**

[1]Department of Drug Design and Pharmacology, University of Copenhagen, Copenhagen, Denmark

[2]Department of Medical and Surgical Sciences (DIMEC), University of Bologna, Bologna, Italy

[†] Equal contribution to first authorship

*** Correspondence:**
Angela Boccia
Department of Drug Design and Pharmacology
University of Copenhagen
Universitetsparken 2, 2100 Copenhagen, Denmark
Email: angela.boccia4@unibo.it

**Abstract**

**Introduction**: Rising volumes of individual case safety reports (ICSRs) in spontaneous reporting systems have challenged the capacity of manual, expert-driven individual-level causality assessment in pharmacovigilance. Existing approaches for intelligent automation of this task have primarily relied on single large language models (LLMs), potentially constraining performance on reasoning-intensive components of individual-level causality assessment requiring specialization. The aims of this study were to leverage agentic artificial intelligence (AI) to: (1) develop a modular multi-agent framework for Naranjo-based causality assessment named MARCA (Multi-Agent Retrieval-Augmented Causality Assessment) and (2) evaluate agreement between MARCA-assisted and human expert assessments and explore differences between non-retrieval-augmented generation (RAG) and RAG-labelled question-level scores in a subset of ICSRs.

**Methods**: MARCA was developed as an orchestrated framework including four specialized agents for scientific literature retrieval, drug label assessment, objective evidence assessment, and clinical causality reasoning, respectively. The agents were powered by Gemini 2.5 Pro and DeepSeek-V3.2 and complemented by RAG, which was developed using expert-evaluated ICSRs. Human expert causality assessments were performed for 215 ICSRs comprising 1,562 drug-event combinations (DECs), stratified-sampled from the FDA Adverse Event Reporting System (FAERS, Q1–Q2 2025). ICSRs assessed by human experts were used as the gold standard reference set for comparing the performance of MARCA. Agreement with the gold standard was assessed using percentage agreement and Cohen's κ.

**Results:** In 1,562 DECs, agreement between MARCA-assisted and expert reference scores was 96.8% for Naranjo Question 2 (κ = 0.491; 95% Confidence Interval 0.265-0.663), 85.4% for Question 5 (κ = 0.269; 95%CI 0.180-0.340), and 96.8% for Question 10 (κ = 0.922; 95% CI 0.885-0.951). Final causality classifications agreed in 79.4% (95% CI 71.8-85.3) of combinations (κ = 0.587; 95%CI 0.450-0.690). In the subset where RAG was used, stored RAG-labelled scores showed higher agreement than non-RAG scores for Questions 5 (65.0% vs 83.3%) and 10 (18.3 vs 100%).

**Discussion:** MARCA achieved high percentage agreement with expert when performing causality assessments using the Naranjo algorithm on ICSRs retrieved from FAERS.

# 1 Introduction

In the United States (US), the FDA Adverse Event Reporting System (FAERS) received more than two million individual case safety reports (ICSRs) of suspected post-authorization adverse events (AEs) in 2025, nearly three times the volume reported in 2011 [1]. Individual-level causality assessment is one of the most time- and resource-intensive tasks in pharmacovigilance (PV) and has become increasingly challenging as reporting volumes continue to grow [2, 3].

Off-the-shelf, general-purpose large language models (LLMs) have been investigated as potential tools to support individual-level causality assessment. Abate et al. evaluated general-purpose LLMs using the World Health Organization (WHO) causality assessment tool. Their findings revealed difficulties in identifying listed AEs in the Summary of Product Characteristics (SmPC), along with inconsistent clinical and pharmacological reasoning [4]. Subsequently, Heckmann et al. evaluated biomedical LLMs using structured prompt engineering and the Naranjo algorithm, finding the optimal prompt-algorithm combination and achieving 63.3% agreement in final causality classifications (Gwet's AC1 = 0.644) [5]. Kefala et al. demonstrated that a state-of-the-art general-purpose LLM could accurately identify listed AEs directly from product labels when given a targeted task, addressing a specific limitation that had constrained the assessment of the first Naranjo question in previous experiments [6]. More recently, Heckmann et al. applied inference-time temperature optimization, increasing agreement in final individual-level causality classifications between human experts and LLMs from a baseline of 45.0% to 72.0%, highlighting the substantial influence of inference settings on LLM performance in causality assessment [7]. Despite these advances, previous experiments have demonstrated persistent limitations in addressing pivotal questions of the Naranjo algorithm that require multistep clinical judgment, particularly those involving temporal plausibility, the identification of alternative causes of AEs, and the assessment of objective evidence.

Multi-agent architecture offers a potentially suitable approach to addressing these limitations by decomposing complex causality assessments tasks into specialized subtasks, allowing individual agents to focus on distinct components of the assessment process in the subtask [8]. Furthermore, previous experiments did not explicitly incorporate expert-derived knowledge into LLM-based systems. Retrieval-augmented generation (RAG) was therefore hypothesized to support specialized tasks requiring clinical judgment by providing agents with access to external, expert-curated information rather than relying solely on knowledge acquired during model pretraining [9].

To our knowledge, no published study has combined a multi-agent architecture with selective RAG for individual-level causality assessment in PV. Therefore, this study aimed to (1) develop a Multi-Agent Retrieval-Augmented Causality Assessment (MARCA) framework for Naranjo-based causality assessment and (2) quantify agreement between MARCA-assisted assessments and a human expert reference standard and explore the question-specific differences between stored non-RAG and RAG-labelled assessments.

## 2 Materials and Methods

### 2.1 Study Design

An overview of the study design is provided in Figure 1. MARCA assessed preprocessed FAERS ICSRs through four specialized LLM agents responsible for scientific evidence retrieval, drug label assessment, objective evidence assessment, and clinical causality assessment. A CrewAI orchestrator coordinated agent execution and information exchange. Selective RAG, supported by an expert-curated knowledge base, was evaluated for Naranjo questions requiring complex clinical judgment, specifically temporal plausibility (Question 2, *Q2*), alternative causes (Q5), and objective evidence (Q10). The framework generated question-level rationales and causality assessments, which were evaluated against human expert assessments and benchmarked descriptively against previously published single-agent LLM approaches.

### 2.2 Data Sources, Pre-Processing, and Sampling

A total of 2,577,782 records were extracted from FAERS quarterly data files covering the first two quarters (Q1–Q2) of 2025. Duplicate records were handled according to the predefined data-cleaning procedure, retaining the most recent or most complete record for each PRIMARYID [10]. ICSRs were integrated using their unique report identifiers (PRIMARYID) to reconstruct case-level records containing patient demographics, drug exposures, reported AEs, clinical outcomes, therapeutic indications, and reporter characteristics (Supplementary Figure 1 and Supplementary Table 1) [11, 12]. Patient ages were standardized to years using the corresponding FAERS age-unit codes, and implausible values were excluded. Drug names were normalized to RxNorm concepts (RxCUIs) using the Unified Medical Language System (UMLS) 2025AA release, while AE terms were coded to MedDRA version 27.1 Preferred Terms (PTs) were mapped to their corresponding High-Level Group Terms (HLGTs) and System Organ Classes (SOCs). To obtain a computationally feasible dataset, a sampling procedure was implemented (Supplementary Figure 2). Drug-event combinations (DECs) were constructed by pairing reported drug entries with adverse event terms within each ICSR. Drug role codes recorded in FAERS were retained to distinguish primary suspect (PS), secondary suspect (SS), and concomitant (C) medications. Drug active ingredients were subsequently mapped to Anatomical Therapeutic Chemical (ATC) classification level 4, while AE terms were mapped to MedDRA HLGTs. ATC level 4-HLGT combinations were then randomly sampled using the `slice_sample()` function from the dplyr package (version 1.2.1) in R (version 4.5.1), with a fixed random seed (1234) to ensure reproducibility. The sampling procedure preserved coverage across all 1,061 ATC level 4 classes and 337 HLGT categories represented in the source dataset, which comprised 55,177 FAERS cases, 4,668 unique drugs, and 9,168 distinct PTs. From the stratified dataset, 215 ICSRs were selected for expert causality assessment based on feasibility. These reports yielded 1,562 DECs, which constituted the evaluation dataset. ICSRs containing the sampled ATC level 4–HLGT combinations were subsequently retrieved, and structured clinical narratives were generated programmatically using a standardized template incorporating patient demographics, drug exposures, therapeutic indications, reported AEs, dosing information, clinical outcomes, and treatment timelines. These structured narratives served as the primary case inputs for both human expert assessment and the MARCA framework (Supplementary Figure 3).

### 2.3 Expert Causality Assessment

Causality was assessed using the Naranjo Adverse Drug Reaction Probability Scale [13]. The junior assessor evaluated 215 ICSRs, encompassing 1,562 DECs. For each Naranjo question, the junior assessor assigned a numerical score and documented the supporting clinical rationale in a structured assessment dataset. All assessments were subsequently reviewed by a senior pharmacovigilance expert, and question-level scores were reconciled to establish the final consensus reference standard. These consensus assessments constituted the human expert gold standard against which MARCA outputs were evaluated. The assessment criteria, Standard Operative Procedures, and review procedures are described in Appendix 1.

## 2.4 MARCA System Architecture

MARCA was developed using the CrewAI multi-agent orchestration framework (version 0.126.0), comprising four specialized agents described in sections 2.4.1-2.4.4 [8, 14]. The Naranjo algorithm and chain-of-thought (CoT) prompting were selected as the causality assessment method and prompting strategy, respectively, based on their superior performance in previous comparative studies [5, 7, 15]. All models were configured with a sampling temperature of zero to minimize output variability and enhance reproducibility [7]. The overall architecture and representative prompts are presented in Figure 1 and Supplementary Table 2.

### *2.4.1 Agent 1 - Scientific Evidence Retrieval*

Agent 1, powered by Gemini 2.5 Pro, was designed to retrieve and synthesize scientific evidence relevant to each DEC. For each DEC, PubMed was queried through the National Center for Biotechnology Information (NCBI) Entrez E-utilities Application Programming Interface (API) using a structured search string combining the drug, AE term, and the expression "adverse event" (i.e., "[drug name] AND [adverse event term] adverse event"). Up to three article records were retrieved, including their PubMed identifiers (PMIDs), titles, and available abstracts. The retrieved information was formatted into a standardized literature corpus and provided to a CrewAI agent instructed to summarize the drug's mechanism of action (MoA) and the available clinical evidence relevant to the reported DEC. The resulting narrative summary was subsequently passed to Agent 4 as supporting evidence for causality assessment. The input structure, processing procedure, prompt template, and expected output are illustrated in Supplementary Figure 4.

### *2.4.2 Agent 2 - Drug Label Assessment*

Agent 2, powered by DeepSeek-V3.2, was developed to assess the listedness of reported AEs in FDA-approved drug labeling and provide supporting information for Naranjo Question 1. For each drug, a reference file containing AE terms was generated from the product label using the approach described by Kefala et al. [6]. This procedure involved extracting the relevant safety information from the product label and organizing the identified AEs into a drug-specific reference list. For each DEC, the reported AE term was compared with the entries in the corresponding drug-label reference file. When an exact match was identified after case normalization, the event was classified as "Listed in label." When no matching term was identified, the LLM was prompted to evaluate whether the reported AE was medically related to the AE in the product label based on clinical reasoning. The model returned a categorical assessment (Yes, No, or Unclear) together with a brief clinical explanation. The input structure, decision rules, prompt template, and output categories are illustrated in Supplementary Figure 5.

### *2.4.3 Agent 3 - Objective Evidence Assessment*

Agent 3, powered by DeepSeek-V3.2, was designed to support the assessment of Naranjo Question 10 by distinguishing objectively observable or measurable clinical signs from subjective symptoms. For each reported AE term, the agent applied a structured prompting strategy instructing the model to classify the event as either a "Sign" (value = 1) or a "Symptom" (value = 0), based on whether the event represented an objectively observable clinical finding or a subjective patient-reported experience. Psychiatric disorders were explicitly assigned to the symptom category according to the implemented classification rules. The agent generated a structured JSON output containing the AE term, its classification, the corresponding numerical value, and a brief medical rationale supporting the decision. This classification provided a standardized, term-level proxy for objective evidence relevant to Naranjo Question 10. The input structure, classification rules, prompt template, and expected output are presented in Supplementary Figure 6.

#### *2.4.4 Agent 4 - Clinical Causality Assessment*

Agent 4, powered by Gemini 2.5 Pro, was designed to integrate case-specific clinical information with the scientific evidence synthesized by Agent 1 to generate a comprehensive causality assessment for each DEC. The input comprised a structured clinical context containing available case demographics, relevant medical history, drug exposure, therapeutic indications, treatment dates, adverse event timing, dosage information, dechallenge and rechallenge findings, clinical outcomes, and concomitant medications. Within the CrewAI workflow, Agents 1 and 4 formed a sequential dependency, whereby the literature summary generated by Agent 1 was made available to Agent 4 as additional context. Agent 4 was instructed to produce a standardized causality report addressing temporal plausibility, biological plausibility, dechallenge and rechallenge, objective evidence, confounding factors, previous similar reactions, and an overall synthesis. The report included clinical reasoning relevant to Naranjo questions and a descriptive plausibility rating (High, Medium, or Low). All reasoning was required to be based on the available case information and supporting literature. In the absence of retrieved scientific evidence, the assessment could rely on the available structured case information, without assuming that missing clinical information was negative evidence. The input structure, prompt template, sequential integration with Agent 1, and expected output format are illustrated in Supplementary Figure 7.

#### *2.4.5 Output of MARCA*

The four agents generated complementary drug-label, objective sign-symptom, literature, and clinical reasoning outputs at DEC level. The human junior assessor then reviewed the agent-generated rationales and assigned numerical Naranjo scores, which were stored separately from the consensus human expert reference scores. Accordingly, the evaluated workflow incorporated human scoring of agent-generated evidence rather than fully automated question-level scoring. The resulting question-level scores and supporting rationales were recorded alongside the corresponding consensus human expert assessments in a structured dataset. Total Naranjo scores were calculated separately for the MARCA-assisted and human expert assessments by summing the scores across all ten Naranjo questions. Final causality classifications were then derived using the standard Naranjo thresholds (Definite, Probable, Possible, and Doubtful).

### 2.5 Selective Retrieval-Augmented Generation

RAG was selectively applied to Naranjo Questions 2 (temporal relationship), 5 (alternative causes), and 10 (objective evidence), which had been identified as particularly challenging in previous LLM-based causality assessments [5, 7]. For each DEC, an initial assessment was generated without RAG using the available case information. When the RAG module was enabled, it generated both non-RAG and RAG-assisted question-level outputs, without using the human expert score as a prerequisite for retrieval. The RAG-assisted outputs incorporated expert-annotated reference cases, question-specific scoring rules, and relevant clinical rationales retrieved from a knowledge base (KB). To prevent direct self-retrieval, all records belonging to the ICSR under evaluation were excluded from the retrieval pool using the corresponding FAERS case identifier. Drug pharmacological classes were resolved sequentially using the openFDA drug label API, the RxNorm API, and, when neither source provided a suitable classification, through DeepSeek-V3.2. Three term frequency–inverse document frequency (TF-IDF) indices were constructed to support hierarchical retrieval at the individual-drug, pharmacological-class, and case levels. Candidate reference cases were ranked according to cosine similarity between their TF-IDF representations and the corresponding query. Retrieval strategies were tailored to the clinical requirements of each Naranjo question. For Question 2, retrieval prioritized reference cases involving the same drug, expanding to pharmacologically related drugs when fewer than three suitable references were identified, with global retrieval used as a fallback. When both treatment initiation and AE onset dates were unavailable, retrieval was restricted to reference cases with similarly missing temporal information, and a predefined score of zero was incorporated into the prompt. For Question 5, the same hierarchical retrieval strategy was applied, with preference given to reference cases exhibiting comparable information availability regarding concomitant medications and comorbidities. For Question 10, retrieval relied exclusively on AE term similarity, independently of drug identity, reflecting the agent's task of distinguishing objectively observable clinical signs from subjective symptoms. Up to three reference cases were retrieved for Questions 2 and 5, and up to two for Question 10. A minimum cosine similarity threshold of 0.01 (arbitrary) was applied during hierarchical retrieval. The retrieved references were incorporated into question-specific prompts for DeepSeek-V3.2 (the template is provided in the GitHub repository), together with their retrieval tier, cosine similarity score, case identifier, drug, AE term, and corresponding human expert score and clinical rationale. The model was instructed to use these references to inform and calibrate its assessment rather than mechanically reproduce the retrieved expert scores. The resulting RAG-assisted outputs were subsequently evaluated against the human expert reference standard. A retrospective ablation analysis was conducted on a subset of 60 DECs for which paired non-RAG and RAG-labelled question-level assessments were available. The non-RAG assessments in this subset exhibited disagreement with the human expert reference standard on at least one of Naranjo Questions 2, 5, or 10. The subset was used to compare question-level agreement between the two conditions and was not assumed to represent all discordant DECs in the complete evaluation dataset. The provenance of the RAG-labelled scores as original retrieval-run predictions could not be independently verified from the saved evaluation exports. All prompts were finalized before the evaluation was performed and remained unchanged throughout the assessment. Structured reasoning prompts were used across the framework, with standardized output formats specified for the drug-label and objective-evidence assessment agents.

### 2.6 Study Outcomes and Acceptable Agreement Thresholds

The primary outcome was the agreement for each DEC between MARCA and human expert assessments at the individual Naranjo question level, evaluated for Questions 2–10 at the DEC level using percentage agreement and Cohen's κ coefficient [16, 17]. Question 1 was excluded from the primary comparative analysis because its assessment relied primarily on FDA drug-label listedness detection through Agent 2, rather than the multistep clinical reasoning targeted by MARCA, and because automated AE listedness detection had previously been evaluated separately [6]. Nevertheless, Question 1 was retained in the calculation of the cumulative Naranjo score and evaluated descriptively. The secondary outcome was agreement in the final causality classification, assessed using percentage agreement and unweighted Cohen's κ. Final classifications were derived by summing the scores across all ten Naranjo questions and applying the standard classification thresholds: Doubtful (≤0), Possible (1–4), Probable (5–8), and Definite (≥9). Exact agreement in cumulative numerical scores was also examined. For questions with invariant scores across both assessments, Cohen's κ was considered non-estimable, and percentage agreement was reported descriptively.
An exploratory outcome evaluated the incremental contribution of selective RAG to question-level agreement for Questions 2, 5, and 10. This retrospective ablation analysis included 60 DECs for which the initial non-RAG assessment disagreed with the human expert reference standard on at least one of the three targeted questions (i.e, Questions 2, 5, and 10). Agreement with the reference standard was compared between non-RAG and RAG-assisted assessments to characterize the effect of retrieval on question-level scoring (Questions 2, 5, and 10).
Predefined question-specific percentage-agreement thresholds, established in consultation with industry collaborators before the MARCA evaluation, were used as performance benchmarks (Supplementary Table 3). Thresholds were set at >95% for Questions 1–4 and 8, reflecting their anticipated reliance on relatively direct evidence from regulatory product information or case-specific clinical data, and >60% for Questions 5, 9, and 10, which were considered more dependent on complex clinical interpretation or incompletely documented information. No predefined thresholds were established for Questions 6 and 7 because placebo-response information and toxic drug concentration measurements are rarely available in FAERS reports.

### 2.7 Statistical Analysis

Continuous variables were summarized using means, medians, and standard deviations, while categorical variables were presented as frequencies and percentages. The characteristics of the human expert-assessed dataset were descriptively compared with those of the broader stratified FAERS dataset to evaluate similarities in patient demographics, drug exposures, and AE distributions. Agreement between MARCA and human expert assessments was evaluated at the individual DEC level using percentage agreement and Cohen's κ coefficient. Question-level agreement was calculated for all ten Naranjo questions, with Question 1 reported descriptively but excluded from the primary comparative analysis, which focused on Questions 2–10. Cumulative scores used for final causality classification were calculated by summing all ten Naranjo questions (Q1–Q10), including Question 1. The resulting scores were classified as Doubtful (≤0), Possible (1–4), Probable (5–8), or Definite (≥9). Agreement in final causality classifications was evaluated using percentage agreement and Cohen's κ. For questions with invariant scores across both assessments, Cohen's κ was considered non-estimable, and percentage agreement was reported descriptively. Agreement patterns were summarized using question-level comparison tables and confusion matrices reporting absolute frequencies and within-expert-category percentages. Directional scoring disagreements were evaluated by calculating the difference between MARCA-generated and human expert-assigned scores for each DEC and Naranjo Question 2–10. Positive differences indicated higher scores assigned by MARCA, whereas negative differences indicated lower scores relative to the human expert reference standard. The frequencies of positive, negative, and concordant assessments were calculated for each question, together with mean signed score differences, to identify systematic directional patterns in causality scoring. MARCA performance was descriptively benchmarked against three previously published LLM-based causality assessment studies [4, 5, 7], using percentage agreement, Cohen's κ or Gwet's AC1 where reported, and available score-error measures. Cross-study comparisons were restricted to outcomes considered sufficiently comparable in terms of causality assessment algorithm, unit of analysis, dataset characteristics, and metric definition. Outcomes that could not be meaningfully compared were designated as not comparable (NC), whereas metrics unavailable in the original publications were designated as not reported (NR). No formal pooled statistical comparisons were performed across studies. The incremental contribution of selective RAG was evaluated using question-level percentage agreement, unweighted Cohen's κ, mean signed score differences, and frequencies of overestimation, agreement, and underestimation were compared between non-RAG and RAG-assisted assessments to characterize the effect of retrieval on scoring agreement and directional disagreement patterns.

## 2.8 Implementation and Reporting Standards

Data acquisition, preprocessing, harmonization, and sampling were performed in R version 4.5.1. MARCA, RAG retrieval, and statistical analyses were implemented in Python version 3.11. TF-IDF vectorization and cosine-similarity retrieval were performed using scikit-learn, while CrewAI was used for multi-agent orchestration. External API integrations include openFDA, RxNorm, NCBI Entrez/PubMed, Google Gemini, and DeepSeek. LLM inference was performed through API-based access to Gemini 2.5 Pro and DeepSeek-V3.2, with temperature set to zero. Case and literature context provided to the models was truncated to 1,400 and 800 characters, respectively, to manage computational and API constraints. All evaluation runs were conducted without prompt modification after the assessment protocol was finalized. Analysis code, agent prompts, and model outputs are available through the GitHub repository provided in the Data Availability statement.

# 3 Results

## 3.1 ICSR and Drug-Event Combination Characteristics

The expert-assessed dataset comprised 215 unique ICSRs, encompassing 1,562 DECs. Female patients accounted for 131 ICSRs (60.9%), while the United States of America was the most frequently reported country of the reporter (130 ICSRs, 60.5%). Patient age was available for 141 ICSRs (65.6%), with a median age of 59 years. The median number of reported AEs per ICSR was 3 (mean 4.35, standard deviation [SD] 4.34; range 1–53). The median number of distinct drugs represented among the evaluated DECs was 1 (mean 1.23, SD 0.80; range 1–7). The number of DECs per ICSR had a median of 4 (mean 7.27, SD 27.18; range 1–371), with one ICSR contributing 371 DECs (23.8% of the evaluated combinations) (Supplementary Tables 4 and 5).

## 3.2 Question-Level Agreement

A representative example of an ICSR evaluated using MARCA is presented in Figure 2. Agreement between MARCA and human expert assessments for Naranjo Questions 2–10 is summarized in Table 1. Although excluded from the primary question-level analysis, Question 1 showed 89.3% agreement with the expert reference standard (unweighted Cohen's $\kappa = 0.787$), below the prespecified >95% agreement threshold. Perfect percentage agreement (100%) was observed for Questions 3, 4, and 6–9. For Question 3, both MARCA and human experts assigned a score of zero to 1,546 DECs and +1 to the remaining 16 DECs (Cohen's $\kappa = 1.000$). In contrast, all DECs received a score of zero for Questions 4 and 6–9, resulting in perfect percentage agreement but non-estimable Cohen's $\kappa$ coefficients due to the absence of score variability. Among questions exhibiting disagreement, Question 10 demonstrated the highest percentage agreement (96.8%) and chance-corrected agreement (Cohen's $\kappa = 0.922$; 95% CI 0.885-0.951), with similar mean scores between MARCA and human expert assessments (0.70 vs. 0.72, respectively). Question 2 also demonstrated high percentage agreement (96.8%), although Cohen's $\kappa$ was lower ($\kappa = 0.491$; 95%CI 0.265-0.663), reflecting the predominance of zero scores in both assessments. Question 5 exhibited the lowest percentage agreement (85.4%) and Cohen's $\kappa$ ($\kappa = 0.269$; 95%CI 0.180-0.340). The mean score for Question 5 was lower for MARCA than for human expert assessments (−0.85 vs. −0.50, respectively), indicating a systematic tendency toward lower causality scores when evaluating alternative causes.

## 3.3 Direction of Scoring Disagreements

The direction and magnitude of scoring disagreements between MARCA and human expert assessments are summarized in Table 2. Score differences were calculated as the MARCA-assigned score minus the corresponding human expert score, with positive and negative values indicating overestimation and underestimation, respectively. Question 5 exhibited the largest directional disagreement, with MARCA assigning lower scores than human experts in 205 of 1,562 DECs and higher scores in 23 DECs, resulting in a mean signed difference of −0.350. For Question 2, MARCA assigned higher scores in 44 DECs and lower scores in 6 DECs, yielding a small positive mean difference (+0.049). For Question 10, MARCA assigned lower scores in 42 DECs and higher scores in 8 DECs, resulting in a mean signed difference of −0.022. Overall, the largest systematic difference was observed for Question 5, indicating that MARCA tended to assign lower causality scores than human experts when evaluating alternative causes. This finding does not independently establish whether more alternative causes were correctly identified.

### 3.4 Classification-Level Agreement

At the final Naranjo classification level, based on cumulative scores across all ten questions (Q1–Q10), MARCA and human expert assessments agreed for 79.4% (95% Confidence Interval, *95%CI* 71.8–85.3) of the 1,562 DECs (unweighted Cohen's κ = 0.587, 95%CI 0.450–0.690). Exact agreement in cumulative numerical scores was 71.6%. The mean total scores were 0.48 for MARCA and 0.74 for human experts, corresponding to a mean signed difference of −0.262. The distribution of MARCA classifications according to the expert-assigned category is presented in Table 3. Agreement was highest for Doubtful: 765 of 867 expert-assessed Doubtful DECs were also classified as Doubtful by MARCA (88.2%); 100 were classified as Possible and two as Probable. For the Possible category, 470 of 685 DECs were classified concordantly (68.6%), while 195 were classified as Doubtful and 20 as Probable by MARCA. For the Probable category, six of ten DECs were classified concordantly (60.0%), while four were classified as Possible. No DECs were classified as Definite by either approach.

### 3.5 Effect of Selective Retrieval-Augmented Generation

Table 4 compares the stored non-RAG and RAG-labelled scores in the retrospective subset of 60 DECs for which at least one initial non-RAG score disagreed with the human expert reference standard on Questions 2, 5, or 10. For Question 10, percentage agreement increased from 18.3% in the non-RAG condition to 100.0% in the RAG-labelled condition, while Cohen's κ increased from 0.007 to 1.000. Under non-RAG conditions, MARCA assigned a lower score than the human expert assessment in 49 of 60 DECs, with a mean signed score difference of −0.817. No disagreements remained in the RAG-labelled scores, and the mean signed difference was 0.000. For Question 5, percentage agreement increased from 65.0% to 83.3%, while Cohen's κ increased from 0.016 to 0.571. The number of lower MARCA scores decreased from 20 to ten of 60 DECs, and the mean signed difference changed from −0.967 to −0.500. For Question 2, percentage agreement increased from 98.3% to 100.0%, with the initial disagreement resolved. Cohen's κ was 0.000 for non-RAG scores and was non-estimable for the RAG-labelled scores because both sets of ratings contained only an invariant category. Because the ablation cohort was retrospectively selected on the basis of disagreement with the expert reference standard, these results describe changes in saved score agreement within this subgroup, rather than overall performance or prospectively validated RAG effectiveness. In addition, RAG-labelled predictions were exported from the comparison workbook rather than matched to independently verified original retrieval-run logs.

### 3.6 Comparison with Previous Studies

Table 5 compares MARCA with three previous studies evaluating LLM-based causality assessment: Abate et al. [4], Heckmann et al. (2026a) [5], and Heckmann et al. (2026b) [7]. Differences in causality algorithms, units of analysis, and dataset composition limit direct comparison across studies; consequently, the comparisons are descriptive and do not establish superiority in a shared test population. Accordingly, comparisons were restricted to reported metrics considered sufficiently comparable. Abate et al. used the WHO causality assessment algorithm, so the question 2 and 5 values shown for that study are approximate conceptual analogues (WHO questions 2.5 and 1) rather than like-for-like Naranjo items, and no WHO analogue exists for question 10.

Among the studies applying the Naranjo algorithm, MARCA showed 96.8%, 85.4%, and 96.8% agreement for Questions 2, 5, and 10, respectively. The corresponding values reported by Heckmann et al. (2026b) were 63.6%, 74.1%, and 65.4%, yielding descriptive differences of 33.2, 11.3, and 31.4 percentage points. Findings from Abate et al., who used a different causality assessment framework, were presented as contextual evidence rather than directly comparable Naranjo question-level estimates.

At the final classification level, MARCA showed 79.4% agreement with expert assessment (Cohen's $\kappa = 0.587$). The 71% agreement for GPT-3.5-turbo and 53% for Gemini 1.5 Pro reported by Abate et al. are median question-level values, not final-classification results, and are therefore not directly comparable. The best-performing biomedical LLM configuration evaluated by Heckmann et al. (2026a) achieved 63.3% agreement; the baseline configuration achieved 45.0% and the optimized configuration achieved 72.0% in Heckmann et al. (2026b). MARCA was evaluated without inference-time temperature optimization. The mean MARCA–expert difference in the total Naranjo score was −0.262, indicating a lower average MARCA score than the expert assessment. In contrast, Heckmann et al. (2026b) reported a mean difference of +0.37.

## 3.7 Reporting checklists

TRIPOD-LLM [18] and CHART [19] reporting checklists are provided in Supplementary tables 6 and 7.

# 4 Discussion

MARCA met the predefined percentage-agreement thresholds for the primary Naranjo questions evaluated, but the interpretation of high agreement requires consideration of imbalanced score distributions. In particular, perfect agreement for Questions 4 and 6–9 reflected invariant zero scores rather than discrimination between different clinical assessments. The modular architecture provided separately reviewable evidence summaries and rationales, potentially supporting a tiered expert-review workflow. However, the recorded MARCA scores were compared retrospectively with expert consensus, and the results do not establish that the system can autonomously replace expert causality assessment.

## 4.1 Comparison with Previous Studies

MARCA's high agreement on question 2 may partly reflect a shared default score, since therapy dates are almost never documented in FAERS narratives; the substantially lower Cohen's κ (0.491) indicates that the high percentage agreement is influenced by the predominance of zero scores and should not be interpreted as strong discrimination across the less frequent nonzero categories. Heckmann et al. (2026b) showed the opposite pattern: their model inferred temporal plausibility from narrative context instead of defaulting conservatively, which produced lower percentage agreement (63.6%) and a near-chance κ (0.011) once base-rate imbalance is accounted for [5, 7]. On question 10, MARCA's near-perfect κ substantially exceeds the κ reported by Heckman et al. [5, 7]. On question 5, the limited chance-corrected agreement shared by MARCA and Heckmann et al. (2026b), despite the two systems using entirely different architectures, suggests this question is a structural difficulty for LLM-based Naranjo assessment of spontaneous reports in general, driven by the question's high base-rate skew rather than a weakness specific to either system.

### 4.2 Interpretation of Remaining Disagreement

The remaining disagreements were concentrated in Question 5, which exhibited the lowest agreement between MARCA and human expert assessments. MARCA assigned lower scores more frequently for this question (205 DECs versus 23 with higher scores), resulting in a mean signed difference of −0.350. This pattern indicates a more conservative tendency in scoring alternative causes but does not establish whether this tendency is clinically more accurate. Using the complete Naranjo Q1–Q10 score, discordance in final classifications was concentrated largely at the Doubtful–Possible boundary: 195 expert-assessed Possible DECs were classified as Doubtful by MARCA, whereas 100 expert-assessed Doubtful DECs were classified as Possible. In the 60-DEC exploratory ablation subset, agreement was higher for the saved RAG-labelled scores on Questions 2, 5, and 10. Nevertheless, these analyses were performed in a disagreement-enriched subset, and the RAG-labelled scores could not be independently linked to original retrieval-inference logs. Their differences should not be interpreted as prospective proof of the effect of retrieval.

An additional consideration is the construction of the RAG knowledge base. The KB comprised the expert-assessed DECs used in the evaluation dataset, although the ICSR under evaluation was excluded from its own retrieval pool. Thus, other expert-assessed cases from the evaluation dataset remained available as references. The RAG findings should therefore be interpreted as retrieval-assisted reasoning within the study cohort rather than as performance evaluated against a completely independent knowledge base. Future studies using a strictly held out or externally developed retrieval corpus would provide a stronger test of the generalizability of the RAG component.

### 4.3 Strengths and Limitations

A principal strength of MARCA is its modular architecture, which provides question-specific, reviewable rationales and structured comparisons with human expert assessments. The observed agreement was high on the temporal association and sign-versus-symptom tasks, while evaluation of alternative causes remained more challenging. The improvements in the exploratory comparison of saved non-RAG and RAG-labelled scores require confirmation using independently documented retrieval runs. Several limitations qualify these findings. Spontaneous reports often lack treatment dates, de-challenge and rechallenge outcomes, drug concentrations, and prior reaction histories, limiting the information available for several Naranjo questions. The evaluation comprised 215 ICSRs and 1,562 DECs from one reporting system; individual ICSRs contributed multiple DECs, including one case contributing a substantial proportion of the records, which limits the independence of DEC-level observations. The retrospective RAG ablation involved 60 DECs selected because the non-RAG scores disagreed with expert assessments, and the retrieval knowledge base overlapped with the evaluation cohort, although the assessed ICSR was excluded from its own reference pool. The provenance of the saved RAG-labelled scores was not independently verified against original inference logs. Finally, the framework used one combination of foundation models (Gemini 2.5 Pro and DeepSeek-V3.2), so the contribution of the architecture cannot be separated from model choice. An additional limitation is the clustering of multiple drug–event combinations within individual ICSRs. Because each DEC was treated as a separate observation in the primary agreement analyses, individual reports contributing numerous combinations may have disproportionately influenced the overall estimates.

## 4.4 Future Directions

A continuously growing, human-in-the-loop KB is a natural next step, with newly emerging safety signals vetted by an expert before being added to the retrieval corpus, while maintaining expert review and clear separation between retrieval and evaluation data. Benchmarking against a larger panel of independent expert assessors, using majority scoring as the reference standard, would also reduce reliance on any single assessor's judgement. Evaluating MARCA using data from other spontaneous reporting systems, including the UK's Yellow Card Scheme, the European Union's EudraVigilance, and the Japanese Adverse Drug Event Report (JADER) database, would help assess its generalizability across different pharmacovigilance settings. Furthermore, evaluating alternative foundation models within the same multi-agent architecture could help distinguish the contribution of task specialization and agent orchestration from that of the underlying model selection.

# 5 Conclusion

MARCA demonstrates the feasibility of organizing individual-level causality assessment into specialized, reviewable components supported by retrieval of scientific and expert-derived evidence. Agreement with expert assessments was high for temporal association and objective sign-versus-symptom classification, while the evaluation of alternative causes remained the principal source of disagreement. The retrospective ablation suggests potential benefits of retrieval, but independent confirmation of retrieval-run provenance and prospective validation are needed. MARCA should therefore be considered a decision-support framework that complements, rather than replaces, expert clinical judgment.

# 6 Conflict of Interest

The authors declare that the research was conducted in the absence of any commercial or financial relationships that could be construed as a potential conflict of interest.

# 7 Author Contributions

DRL: Writing - original draft, Formal analysis, Validation and Visualization. AB: Writing - review and editing, Data curation, Validation, and Visualization. MS: Conceptualization, Formal analysis, Project administration, and Writing – review and editing. All authors contributed to the work, critically reviewed the manuscript, approved the final version, and agree to be accountable for the content of the work.

# 8 Funding & ethics

This research received no specific grant from any funding agency in the public, commercial, or not-for-profit sectors. This study analysed publicly available, de-identified data from the FDA Adverse Event Reporting System. No patients were contacted, and ethical approval and informed consent were not required.

# 9 Acknowledgements

Generative AI was used to support the conceptual design, generation, and refinement of schematic figures included in this manuscript. Figures were developed using ChatGPT (OpenAI, GPT-6 for text-based assistance; accessed 23rd October 2026) and its integrated image-generation functionality. The specific version of the underlying image-generation model was not available in the generation records.

AI assistance included the development of visual layouts, graphical elements, and scientific annotations. All AI-generated figures were critically reviewed, revised, and validated by the authors against the study methodology, source data, and computational framework. The authors take full responsibility for the accuracy, scientific integrity, and final content of all figures.

## 10 Data Availability Statement

The datasets generated for this study can be found in the MARCA: Multi-Agent Retrieval-Augmented for Individual-level Causality Assessment in Pharmacovigilance [https://github.com/daisyraelopez/Multi-Agent-Retrieval-Augmented-for-Individual-level-Causality-Assessment-MARCA-in-Pharmacovigilance/tree/main].

## 12 Tables

**Table 1. Agreement Between MARCA and Human Expert Assessors Across Naranjo Questions.** Agreement between MARCA and human expert assessors across Naranjo Q2–Q10, based on the expert-assessed dataset. Mean MARCA score and mean human score represent the mean scores assigned by each rater for each question. Imbalanced score distributions resulted in reduced or undefined κ values despite high percentage agreement.

| Naranjo Question | Description | Agreement (%) | Cohen's κ (95%CI) | Mean MARCA Score | Mean Expert Score |
|---|---|---|---|---|---|
| **Q2** | Temporal association | 96.8% | 0.491 (0.265-0.663) | 0.09 | 0.04 |
| **Q3** | Improvement following drug discontinuation | 100.0% | 1.000 | 0.01 | 0.01 |
| **Q4** | Reappearance following re-challenge | 100.0% | N/A | 0.00 | 0.00 |
| **Q5** | Alternative cause excluded | 85.4% | 0.269 (0.180-0.340) | −0.85 | −0.50 |
| **Q6** | Reaction following placebo administration | 100.0% | N/A | 0.00 | 0.00 |
| **Q7** | Toxic drug level detected | 100.0% | N/A | 0.00 | 0.00 |
| **Q8** | Dose-response relationship | 100.0% | N/A | 0.00 | 0.00 |
| **Q9** | Similar reaction to previous exposure | 100.0% | N/A | 0.00 | 0.00 |
| **Q10** | Objective evidence confirmation | 96.8% | 0.922 (0.885-0.951) | 0.70 | 0.72 |

**Table 2. Direction of Scoring Disagreement Between MARCA and Human Expert Assessors.** Direction of scoring disagreements between MARCA and human expert assessors across Naranjo questions 2-10. Positive mean differences indicate higher scores assigned by MARCA than by experts, whereas negative values indicate lower scores. A mean difference of zero indicates no directional difference.

| **Naranjo Q** | **MARCA > Expert (n)** | **MARCA = Expert (n)** | **MARCA < Expert (n)** | **Mean difference** |
|---|---|---|---|---|
| Q2 | 44 | 1,512 | 6 | +0.049 |
| Q3, Q4, Q6–Q9 | 0 | 1,562 | 0 | 0.000 |
| Q5 | 23 | 1,334 | 205 | −0.350 |
| Q10 | 8 | 1,512 | 42 | −0.022 |

**Table 3. Naranjo Classification Agreement Between MARCA and Human Expert Assessors by Classification Category.** Agreement between MARCA and human expert assessors across Naranjo classification categories based on the complete Q1–Q10 Naranjo scoring framework. Agreement was calculated as the proportion of expert-assigned DECs within each category that were correctly classified by MARCA. The most common direction of misclassification is reported for each category. *Legend: N/A = not applicable.*

| Naranjo category | Expert-assigned (n) | Correctly classified by MARCA (n) | Agreement (%) | Most common misclassification |
|---|---|---|---|---|
| Doubtful | 867 | 765 | 88.2% | Possible (n = 100, 11.5%) |
| Possible | 685 | 470 | 68.6% | Doubtful (n = 195, 28.5%) |
| Probable | 10 | 6 | 60.0% | Possible (n = 4, 40.0%) |
| Definite | 0 | N/A | N/A | N/A |

**Table 4. Percentage Agreement and Cohen's κ between MARCA and Human expert assessors Under Non-RAG and RAG Conditions.** Question-level evaluation of MARCA against human expert consensus under non-RAG and RAG conditions.
*Legend: κ = Cohen's Kappa; RAG =Retrieval-Augmented Generation.*

| Question | Non-RAG (%) | RAG (%) | κ (Non-RAG) | κ (RAG) |
|---|---|---|---|---|
| Q2 - Temporal association | 98.33% | 100.00% | 0.000 | N/E |
| Q5 - Alternative causes | 65.00% | 83.33% | 0.016 | 0.571 |
| Q10 - Objective evidence | 18.33% | 100.00% | 0.007 | 1.000 |

**Table 5. Cross-Study Comparison of Agreement Metrics for Automated and Human Causality Assessment.** Comparison of key agreement metrics across studies evaluating causality assessment. The Closest conceptual WHO analogues are Q2 (WHO Question 2.5) and Q5 (WHO Question 1); no direct analogue exists for Q10.

*Legend: ND = not directly comparable; NR = not reported by the original authors. AC1 = Gwet's first-order agreement coefficient, not directly comparable to Cohen's κ. Abate et al. final-classification cells are not comparable because the reported percentages represent median question-level agreement.*

| Metric | Abate et al. (2025) [4] | Heckmann et al. (2026a) [5] | Heckmann et al. (2026b) [7] | MARCA (present study) |
|---|---|---|---|---|
| Model | GPT-3.5-turbo / Gemini 1.5 Pro | Medicine LLaMA-3 8B | GPT-5.2 | Gemini 2.5 Pro + DeepSeek-V3.2 |
| Algorithm | WHO | Naranjo | Naranjo | Naranjo |
| Sample | 150 VAERS cases | 150 cases (140 FAERS + 10 VAERS) | 723 FAERS cases | 215 FAERS cases (1,562 DECs) |
| Q2 agreement | 66.0% / 68.0%* | 54.0% | 63.6% | 96.8% |
| Q5 agreement | 54.7% / 30.0%* | 52.7% | 74.1% | 85.4% |
| Q10 agreement | ND | 63.3% | 65.4% | 96.8% |
| κ (Q2 / Q5 / Q10) | NR | NR | 0.011 / 0.180 / 0.220 | 0.491 / 0.269 / 0.922 |
| Final classification agreement | NC — median question-level agreement: 71% (GPT-3.5) / 53% (Gemini) | 63.3% (Gwet's AC1 = 0.644) | 45.0% baseline → 72.0% optimized | 79.4% (κ = 0.587) |
| Mean score error (LLM − expert) | NR | NR | +0.37 | −0.262 |

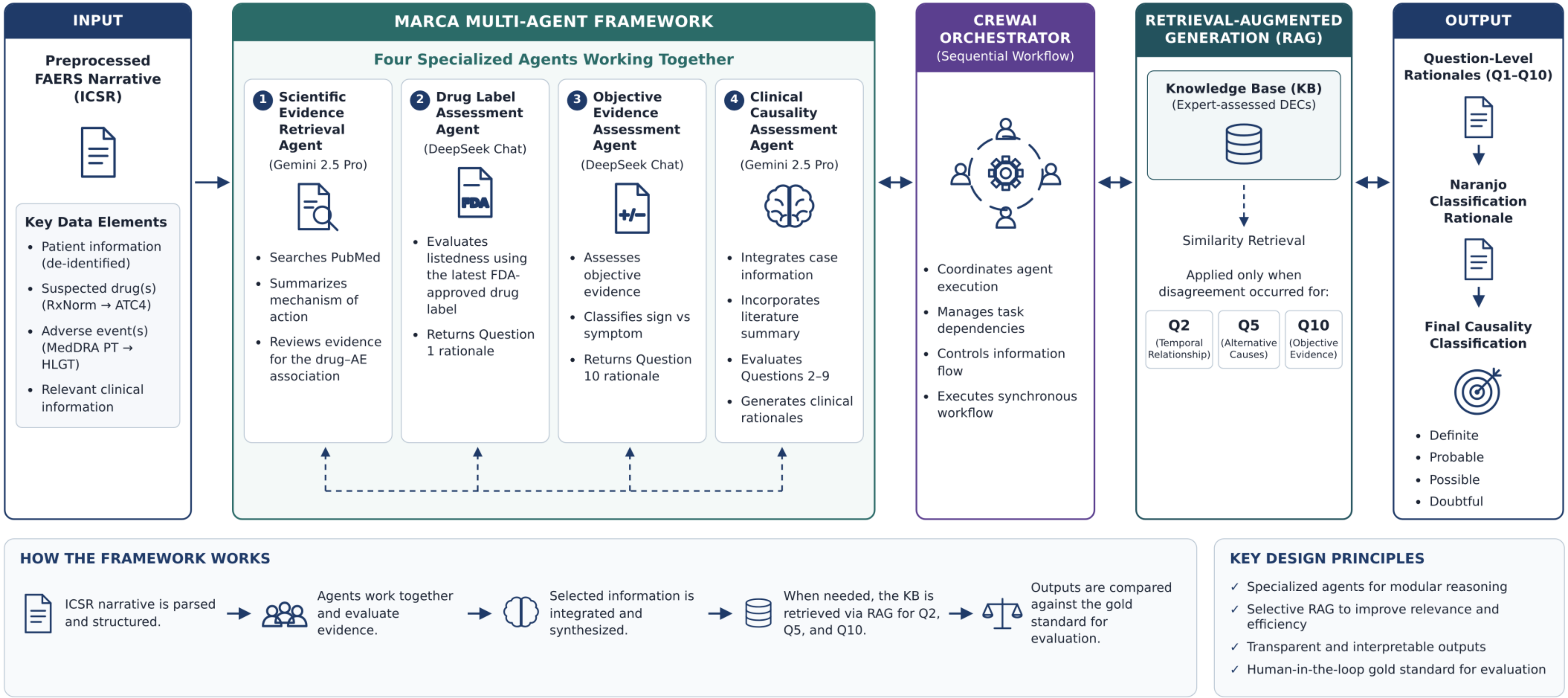

**Figure 1. Study design.** A FAERS individual case safety report is processed by four specialized agents that assess scientific evidence, drug label listedness, objective evidence, and clinical causality. A knowledge base containing expert-assessed drug–event combinations supports selective retrieval-augmented generation for Naranjo Questions 2, 5, and 10. The outputs are integrated to generate question-level rationales, calculate the cumulative Naranjo score, and determine the final causality classification before export for downstream evaluation.

*Legend: AE= adverse event; DEC = drug–event combination; FAERS = FDA Adverse Event Reporting System; FDA = U.S. Food and Drug Administration; ICSR = individual case safety report; KB = knowledge base; LLM = large language model; MARCA = Multi-Agent Retrieval-Augmented Causality Assessment; MOA = mechanism of action; Q = Naranjo Algorithm question; RAG = retrieval-augmented generation.*

*Disclaimer: The figure is AI-generated.*

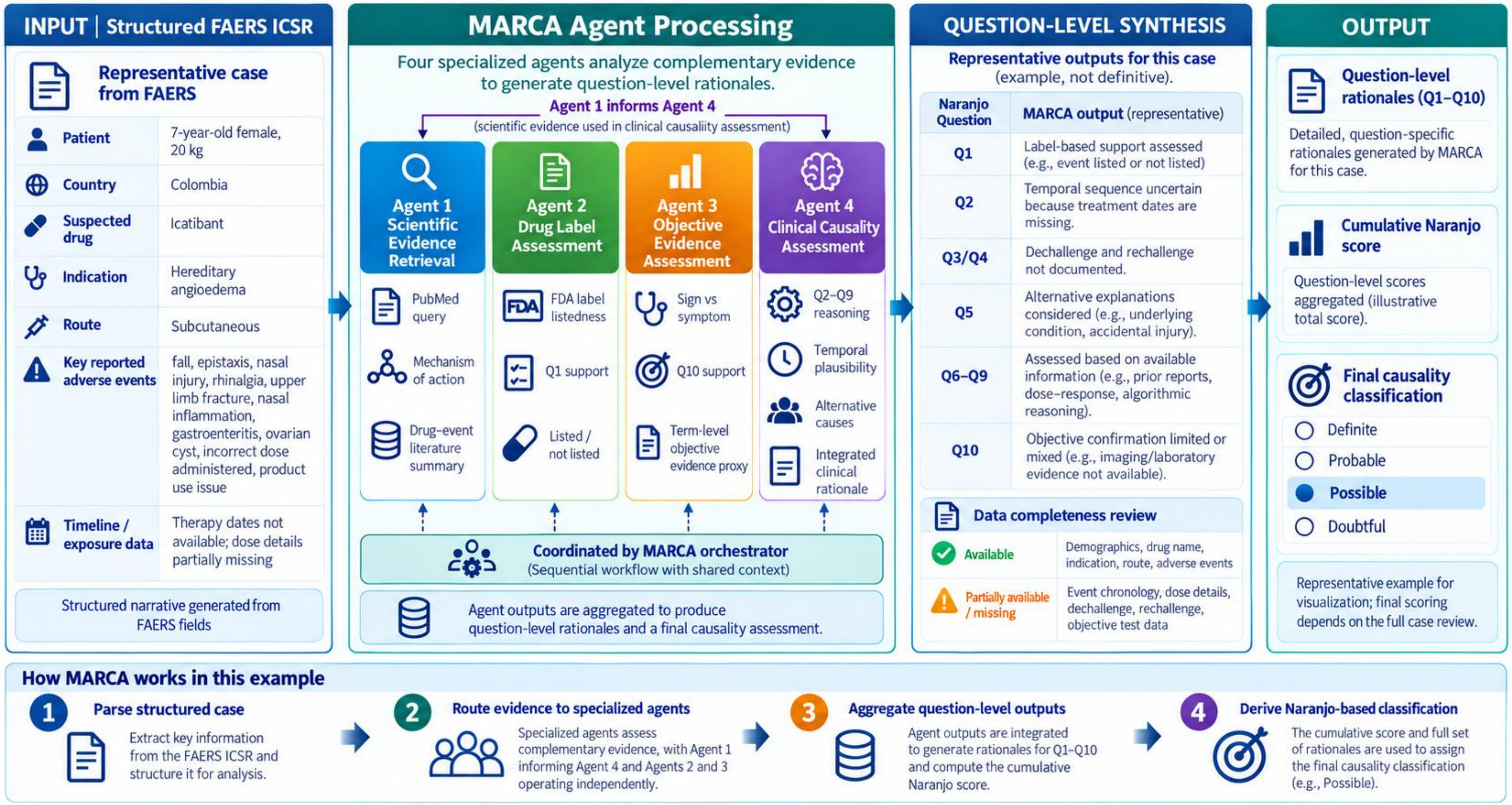

**Figure 2. Representative example of an ICSR evaluated using MARCA.** Example of a structured individual case safety report (ICSR) processed by MARCA, illustrating the integration of patient demographics, suspected drug information, and reported adverse events to assess data completeness, identify missing clinical information, and highlight medication errors and relevant clinical considerations.

*Legend: ICSR = individual case safety report, MARCA = Multi-Agent Retrieval-Augmented Causality Assessment.*

*Disclaimer: The figure is AI-generated.*

# Supplementary Material

## Multi-Agent Retrieval-Augmented Causality Assessment (MARCA) of Individual Case Safety Reports in Pharmacovigilance

**Daisy Rae Lopez [1†], Angela Boccia [1,2†], Maurizio Sessa [1]**

[1]Department of Drug Design and Pharmacology, University of Copenhagen, Copenhagen, Denmark
[2]Department of Medical and Surgical Sciences (DIMEC), University of Bologna, Bologna, Italy
[†] Equal contribution to first authorship

*** Correspondence:**
Angela Boccia
Department of Drug Design and Pharmacology
University of Copenhagen
Universitetsparken 2, 2100 Copenhagen, Denmark
Email: angela.boccia4@unibo.it

**Supplementary Table 1. FAERS ICSR Structured Template.**
*Legend: NA = Not available, CN = Consumer, YR = Year, FAERS = Food and Drug Administration Adverse Event Reporting System*

| **FAERS Variables** |
|---|
| Case ID |
| Case Version |
| Event Date |
| Authorization Number |
| Manufacturer Number |
| Manufacturer Sender |
| Literature Reference |
| Age |
| Age Code |
| Age Group |
| Sex |
| E Sub |
| Weight |
| Weight Code |
| Occupation Code |
| Reporter Country |
| Occurrence Country |
| Adverse Reaction(s) |
| Case ID Number |
| Drug Sequence |
| Role Code |
| Drug Name |
| Product AI |
| Val VBM |
| Route |
| Dose VBM |
| Cumulative Dose (Char) |
| Cumulative Dose (Unit) |
| Dechallenge |
| Rechallenge |
| Lot Number |
| Expiration Date |
| NDA Number |
| Dose Amount |
| Dose Unit |
| Dose Form |
| Dose Frequency |
| Outcome Code |
| FDA Code |
| Manufacture Date |
| Initial FDA Date |
| FDA Date |
| Report Source Code |
| Indication for Use |
| Therapy Start Date |
| Therapy End Date |
| Duration |
| Duration Code |

**Supplementary Table 2. Representative Prompt Designs by Agent.**

| Agent | Representative prompt design |
|---|---|
| **Agent 1 - Scientific Evidence Retrieval (Gemini 2.5 Pro)** | Role: expert medical searcher and literature summarizer. Goal: analyze PubMed results for [drug] and [adverse event] to summarize mechanism of action and clinical evidence. Search string: "[drug name] AND [adverse event term] adverse event" (NCBI Entrez); up to 3 records retrieved per DEC; summary truncated to 800 characters before being passed to Agent 4. |
| **Agent 2 - Drug Label Assessment (DeepSeek-V3.2)** | System prompt: "You are a medical documentation assistant that evaluates AE-drug relations." User prompt (not-listed path only): "Determine if the following adverse event is medically related to the drug based on clinical reasoning. Drug: [drug name]. Adverse Event: [adverse event]. Output: Related – <Yes/No/Unclear>; Explanation – <brief medical explanation>." Events found directly in the FDA label reference file are classified as Listed without invoking this prompt. |
| **Agent 3 - Objective Evidence Assessment (DeepSeek-V3.2)** | System prompt: "You are a clinical pharmacovigilance expert. Classify events as 'Sign' (value 1) or 'Symptom' (value 0)." User prompt: "Analyze the medical term: '[adverse event]'" returns a JSON object with the sign/symptom classification and a brief supporting rationale. |
| **Agent 4 - Clinical Causality Assessment (Gemini 2.5 Pro)** | Receives the full structured case context (demographics, drug exposure, temporal data, dechallenge/rechallenge, concomitant medications; truncated to 1,400 characters) plus the Agent 1 literature summary. Produces a clinical rationale and a descriptive plausibility rating (high/medium/low) for Naranjo questions 2 through 9, following Chain-of-Thought (CoT) reasoning. |

**Supplementary Table 3. Acceptable Agreement Thresholds for each Naranjo Question**. Acceptable agreement thresholds, supporting rationale, and key considerations for each Naranjo Algorithm question. Thresholds were established in consultation with experts from GSK.
*Legend*: DEC = drug-event combination; ICSR = individual case safety report.

| **Naranjo Question** | **Acceptable Agreement** | **Rationale** | **Considerations** |
|---|---|---|---|
| **1** | >95% | Can be determined from the product label. | The term *conclusive* is subjective and may reduce agreement to 60-70%. |
| **2** | >95% | If the event preceded drug exposure, it would not constitute an ICSR. | The key consideration, not addressed by Naranjo, is whether the time to onset is clinically reasonable. |
| **3** | >95% | Can be derived from structured ICSR fields. | For multiple drugs or events, dechallenge may not correspond to the evaluated DEC. Antagonists are rarely reported (primarily naloxone for opioid overdose). |
| **4** | >95% | Can be derived from structured ICSR fields. | Rechallenge may not correspond to the evaluated DEC; antagonists are seldom reported. |
| **5** | >60% | Requires consideration of concomitant drugs, medical history, and patient-related factors. | Co-medications are easier to identify than underlying medical conditions. |
| **6** | N/A | Placebo use does not apply to post-marketing reports. | Not applicable to ICSRs; only theoretical non-exposure periods could serve as a comparator. |
| **7** | N/A | Drug concentrations are rarely reported. | Therapeutic or toxic concentration thresholds are often unavailable except in overdose. |
| **8** | >95% | Can be derived from structured ICSR fields. | Structured fields may not correspond to the same DEC; dose-related severity changes are uncommon. |
| **9** | >60% | Previous exposure is rarely documented and may only appear in free text. | ICSRs are not linked across individuals; information depends on reporter history, and *similar* is drug-class dependent. |
| **10** | >60% | Objective confirmation may only be documented in free text. | The definition of *confirmation* may vary between assessors. |

**Supplementary Table 4. Summary of Continuous Variables for the Total and Expert-assessed Datasets.** Summary statistics for continuous demographic and case-level variables in the total FAERS dataset and the expert-assessed dataset used for MARCA agreement analysis.
*Legend: AE = adverse event; FAERS = FDA Adverse Event Reporting System; SD = standard deviation.*

| Variable | Total Dataset (n = 55,177) | Mean | Median | SD | Expert-assessed Dataset (n = 215) | Mean | Median | SD |
|---|---|---|---|---|---|---|---|---|
| **Age (years)** | 38,227 | 54.90 | 59.0 | 21.16 | 141 | 53.26 | 59.00 | 20.70 |
| **Weight (kg)** | 19,140 | 73.05 | 71.0 | 37.68 | 42 | 65.87 | 67.62 | 25.91 |
| **AEs per case** | 55,133 | 9.38 | 5.0 | 15.20 | 215 | 4.35 | 3.00 | 4.34 |
| **Distinct drugs per case** | 55,133 | 15.80 | 8.0 | 43.67 | 215 | 1.23 | 1.00 | 0.80 |

**Supplementary Table 5. Case-level Characteristics of the Expert-assessed Dataset.** Distribution of DECs, AEs, and drugs per case.
*Legend: AE = adverse event; DEC = drug-event combination.*

| Variable | Min | Median | Max |
|---|---|---|---|
| DECs per case | 1 | 4 | 371 |
| AEs per case | 1 | 3 | 53 |
| Distinct drugs per case | 1 | 1 | 7 |

**Supplementary Table 6. TRIPOD-LLM Checklist**.

*Legend: M = LLM methods; D = de novo LLM development; E = LLM evaluation; H = LLM evaluation in healthcare settings; C = classification; OF = outcome forecasting; QA = long-form question answering; IR = information retrieval; DG = document generation; SS = summarization and simplification; MT = machine translation; API = application programming interface.*

| Section | Item | Reporting guideline item | Design | LLM task | Manuscript location / reporting assessment |
|---|---|---|---|---|---|
| Title | 1 | Identify the study as developing, fine-tuning and/or evaluating the performance of an LLM, specifying the task, the target population and the outcome to be predicted. | All | All | p. 1 (title); p. 2 (abstract)<br>R - Study identification and target task |
| Abstract | 2 | See TRIPOD-LLM for abstracts. | All | All | p. 2 (abstract)<br>R - Structured abstract |
| Introduction | Introduction | Introduction | Introduction | Introduction | |
| Background | 3a | Explain the healthcare context/use case (for example, administrative, diagnostic, therapeutic and clinical workflow) and rationale for developing or evaluating the LLM, including references to existing approaches and models. | All | All | p. 3 (§1)<br>R - Clinical context and rationale |
| Background | 3b | Describe the target population and the intended use of the LLM in the context of the care pathway, including its intended users in current gold standard practices (for example, healthcare professionals, patients, public or administrators). | E<br>H | All | pp. 3, 6, 13 (§§1, 2.4.5, 4.6)<br>P - Intended users not explicitly characterized |
| Objectives | 4 | Specify the study objectives, including whether the study describes the initial development, fine-tuning or validation of an LLM (or multiple stages). | All | All | pp. 2–3 (abstract; §1)<br>R - Study aims |
| Methods | Methods | Methods | Methods | Methods | |
| Data | 5a | Describe the sources of data separately for the training, tuning and/or evaluation datasets and the rationale for using these data (for example, web corpora, clinical research/trial data, EHR data or unknown). | All | All | pp. 4, 7 (§§2.2, 2.5)<br>P - Evaluation and knowledge-base data described; development split not established |
| Data | 5b | Describe the relevant data points and provide a quantitative and qualitative description of their distribution and other relevant descriptors of the dataset (for example, source, languages and countries of origin). | All | All | pp. 4, 9 (§§2.2, 3.1); Supp. Tables 4–5<br>R - Sample characteristics and variables |
| Data | 5c | Specifically state the date of the oldest and newest item of text used in the development process (training, fine-tuning and reward modeling) and the evaluation datasets. | All | All | p. 4 (§2.2)<br>P - Reporting period given; earliest/latest individual record dates not stated |
| Data | 5d | Describe any data preprocessing and quality checking, including whether this was similar across text corpora, institutions and relevant sociodemographic groups. | All | All | p. 4 (§2.2)<br>R - Preprocessing and terminology mapping |
| Data | 5e | Describe how missing and imbalanced data were handled and provide reasons for omitting any data. | All | All | pp. 4, 7, 12–13 (§§2.2, 2.5, 4.4)<br>P - Missingness discussed, but complete missing-data handling not defined |
| Analytical Methods | 6a | Report the LLM name, version and last date of training. | All | All | pp. 5–6, 9 (§§2.4, 2.8)<br>P - Model versions named; last training/release dates absent |
| Analytical Methods | 6b | Report details of the LLM development process, such as LLM architecture, training, fine-tuning procedures and alignment strategy (for example, reinforcement learning and direct preference optimization) and alignment goals (for example, helpfulness, honesty and harmlessness). | M<br>D | All | pp. 5–6 (§2.4)<br>P - Agent architecture described; no novel LLM training or fine-tuning documented |
| Analytical Methods | 6c | Report details of how the text was generated using the LLM, including any prompt engineering (including consistency of outputs), and inference settings (for example, seed, temperature, max token length and penalties), as relevant. | M<br>D<br>E | All | pp. 5–9 (§§2.4–2.5, 2.8); Supp. Table 2<br>P - Prompts/settings summarized, not full inference metadata |
| Analytical Methods | 6d | Specify the initial and postprocessed output of the LLM (for example, probabilities, classification and unstructured text). | All | All | pp. 5–7 (§§2.4.1–2.5)<br>R - Agent outputs, human scoring and RAG scores |
| Analytical Methods | 6e | Provide details and rationale for any classification and, if applicable, how the probabilities were determined and thresholds identified. | All | C<br>OF | pp. 6, 8 (§§2.4.5, 2.6)<br>R - Naranjo rules and classification thresholds |
| LLM Output | 7a | Include metrics that capture the quality of generative outputs, such as consistency, relevance, accuracy and presence/type of errors compared to gold standards. | All | QA<br>IR<br>DG<br>SS<br>MT | pp. 10–11 (§§3.2–3.5); Tables 1–4<br>P - Agreement evaluated; output rationale quality not independently assessed |
| LLM Output | 7b | Report the outcome metrics' relevance to the downstream task at deployment time and, where applicable, the correlation of metric to human evaluation of the text for the intended use. | E<br>H | All | pp. 8, 13 (§§2.6, 4.6)<br>P - Clinical relevance addressed; no deployment outcomes |
| LLM Output | 7c | Clearly define the outcome, how the LLM predictions were calculated (for example, formula, code, object and API), the date of inference for closed-source LLMs and evaluation metrics. | E<br>H | All | pp. 6–9 (§§2.4.5–2.7)<br>P - Metrics/score calculation described; closed-model inference date not reported |
| LLM Output | 7d | If outcome assessment requires subjective interpretation, describe the qualifications of the assessors, any instructions provided, relevant information on demographics of the assessors and inter-assessor agreement. | All | All | p. 5 (§2.3); p. 6 (§2.4.5)<br>P - Assessor roles reported; experience and initial inter-rater agreement absent |
| LLM Output | 7e | Specify how performance was compared to other LLMs, humans and other benchmarks or standards. | All | All | pp. 8–9, 11 (§§2.7, 3.6); Table 5<br>R - Cross-study benchmarking approach |
| Annotation | 8a | If annotation was done, report how the text was labeled, including providing specific annotation guidelines with examples. | All | All | p. 5 (§2.3); Appendix 1<br>P - Structured scoring described; item-specific annotation details in Appendix |
| Annotation | 8b | If annotation was done, report how many annotators labeled the dataset(s), including the proportion of data in each dataset that was annotated by more than one annotator, and the inter-annotator agreement. | All | All | p. 5 (§2.3)<br>P - Junior/senior review described; independent dual-annotation proportion absent |
| Annotation | 8c | If annotation was done, provide information on the background and experience of the annotators or the characteristics of any models involved in labeling. | All | All | p. 5 (§2.3)<br>P - Assessor positions given, not detailed experience |
| Prompting | 9a | If research involved prompting LLMs, provide details on the processes used during prompt design, curation and selection. | All | All | pp. 5–7 (§§2.4–2.5)<br>P - Prompt choice and freeze reported; development/evolution limited |
| | 9b | If research involved prompting LLMs, report what data were used to develop the prompts. | All | All | pp. 5–7 (§§2.4–2.5); Supp. Table 2<br>P - Prompt inputs given; prompt-development dataset not separately described |
| Summariza-tion | 10 | Describe any preprocessing of the data before summarization. | All | SS | pp. 5, 9 (§§2.4.1, 2.8)<br>R - Abstract retrieval and text truncation |
| Instruction tuning/ alignment | 11 | If instruction tuning/alignment strategies were used, what were the instructions, data and interface used for evaluation, and what were the characteristics of the populations doing the evaluation? | M<br>D | All | NA - No instruction fine-tuning reported |
| Compute | 12 | Report compute, or proxies thereof (for example, time on what and how many machines, cost on what and how many machines, inference time, floating-point operations per second), required to carry out methods. | M<br>D<br>E | All | p. 9 (§2.8)<br>P - Constraints mentioned; no compute time, hardware or cost |
| Ethical approval | 13 | Name the institutional research board or ethics committee that approved the study and describe the participant-informed consent or the ethics committee waiver of informed consent. | All | All | NR - No ethics committee approval or waiver statement |
| Open science | 14a | Give the source of funding and the role of the funders for the present study. | All | All | p. 14 (§7)<br>R - No specific funding stated |
| Open science | 14b | Declare any conflicts of interest and financial disclosures for all authors. | All | All | p. 14 (§5)<br>R - Conflict-of-interest statement |
| Open science | 14c | Indicate where the study protocol can be accessed or state that a protocol was not prepared. | H | All | NR - No protocol availability or no-protocol declaration |
| Open science | 14d | Provide registration information for the study, including register name and registration number, or state that the study was not registered. | H | All | NR - No registration status statement |
| Open science | 14e | Provide details of the availability of the study data. | All | All | p. 14 (§8)<br>R - Repository given for study datasets |
| Open science | 14f | Provide details of the availability of the code to reproduce the study results. | All | All | pp. 9, 14 (§§2.8, 8)<br>R - Repository given for analysis code |
| Public involvement | 15 | Provide details of any patient and public involvement during the design, conduct, reporting, interpretation or dissemination of the study or state no involvement. | H | All | NR - Patient/public involvement not addressed |
| Results | Results | Results | Results | Results | |
| Participants | 16a | When using patient/EHR data, describe the flow of text/EHR/patient data through the study, including the number of documents/questions/participants with and without the outcome/label and follow-up time as applicable. | E<br>H | All | pp. 4, 9 (§§2.2, 3.1); Supp. Fig. 2<br>P - Main counts reported; complete selection flow not fully described |
| Participants | 16b | When using patient/EHR data, report the characteristics overall and for each data source or setting and development/evaluation splits, including the key dates, key characteristics and sample size. | E<br>H | All | pp. 4, 9 (§§2.2, 3.1); Supp. Tables 4–5<br>R - Sample and case-level characteristics |
| Participants | 16c | For LLM evaluation that includes clinical outcomes, show a comparison of the distribution of important clinical variables that may be associated with the outcome between development and evaluation data, if available. | E<br>H | All | p. 9 (§3.1); Supp. Table 4<br>P - Descriptive comparison, no development/evaluation split |
| Participants | 16d | When using patient/EHR data, specify the number of participants and outcome events in each analysis (for example, for LLM development, hyperparameter tuning and LLM evaluation). | E<br>H | All | pp. 9–11 (§§3.1, 3.5)<br>R - 215 ICSRs; 1,562 DECs; 60-DEC subset |
| Performance | 17 | Report LLM performance according to prespecified metrics (see item 7a) and/or human evaluation (see item 7d). | All | All | pp. 10–11 (§§3.2–3.5); Tables 1–4<br>R - Primary, secondary, exploratory outcomes |
| LLM updating | 18 | If applicable, report the results from any LLM updating, including the updated LLM and subsequent performance. | All | All | NA - No post-training model update evaluated |
| Discussion | Discussion | Discussion | Discussion | Discussion | |
| Interpretation | 19a | Give an overall interpretation of the main results, including issues of fairness in the context of the objectives and previous studies. | All | All | pp. 11–12 (§§4.1–4.3)<br>P - Results interpreted; fairness not formally analyzed |
| Limitations | 19b | Discuss any limitations of the study and their effects on any biases, statistical uncertainty and generalizability. | All | All | p. 13 (§4.4)<br>R - Limitations described |
| Usability of the LLM in context | 19c | Describe any known challenges in using data for the specified task and domain context with reference to representation, missingness, harmonization and bias. | E<br>H | All | pp. 12–13 (§§4.2–4.4)<br>R - Missing data, data-source limitations and clustering |
| Usability of the LLM in context | 19d | Define the intended use for the implementation under evaluation, including the intended input, end-user and level of autonomy/human oversight. | E<br>H | All | pp. 6, 13 (§§2.4.5, 4.6)<br>P - Human oversight described; intended users not fully specified |
| Usability of the LLM in context | 19e | If applicable, describe how poor quality or unavailable input data should be assessed and handled when implementing the LLM; that is, what is the usability of the LLM in the context of current clinical care. | E<br>H | All | pp. 12–13 (§§4.2, 4.4)<br>P - Incomplete reporting discussed, not an operational deployment procedure |
| Usability of the LLM in context | 19f | If applicable, specify whether users will be required to interact in the handling of the input data or use of the LLM, and what level of expertise is required of users. | E<br>H | All | pp. 6, 13 (§§2.4.5, 4.5–4.6)<br>P - Human intervention described; required expertise not specified |
| Usability of the LLM in context | 19g | Discuss any next steps for future research, with a specific view of the applicability and generalizability of the LLM. | All | All | p. 13 (§4.5)<br>R - Next steps and external evaluation |

## Supplementary Table 7. CHART Checklist

| Section | Item | CHART checklist item | Manuscript location / reporting assessment |
|---|---|---|---|
| Title and Abstract | Title and Abstract | Title and Abstract | |
| Title | 1a | State that the study is assessing one or more generative AI-driven chatbots for clinical evidence or health advice. | pp. 1–2 (title; abstract)<br>R - Clinical causality assessment and agents |
| Abstract/ Summary | 1b | Apply a structured format, if applicable. | p. 2 (abstract)<br>R - Structured abstract |
| Introduction | Introduction | Introduction | |
| Background | 2a | State the scientific background, rationale, and healthcare context for evaluating the generative AI-driven chatbot(s), referencing relevant literature when applicable. | p. 3 (§1)<br>R - Prior work and healthcare rationale |
| Background | 2b | State the aims and research questions including the target audience, intervention, comparator(s), and outcome(s). | pp. 2–3 (abstract; §1)<br>R - Aims, evaluation and comparator |
| Methods | Methods | Methods | |
| Model Identifiers | 3a | State the name and version identifier(s) of the generative AI model(s) and chatbot(s) under evaluation, as well as their date of release or last update. | pp. 5–6, 9 (§§2.4, 2.8)<br>P - Model IDs given; release/last-update dates not stated |
| Model Identifiers | 3b | State whether the generative AI model(s) and chatbot(s) are open-source or closed-source/proprietary. | pp. 5–6 (§2.4)<br>P - Named commercial models; licensing/openness not explicit |
| Model Details | 4a | State whether the generative AI model was a base model or a novel base model, tuned model, or fine-tuned model. | pp. 5–6 (§2.4)<br>P - Uses named base models, but tuning status not explicit |
| Model Details | 4b | If a base model is used, cite its development in sufficient detail to identify the model. | pp. 5–6 (§2.4)<br>P - Model names given; model-development sources not fully cited |
| Model Details | 4c | If a novel base model, tuned model, or fine-tuned model is used, describe the pre- and/or post-implementation/deployment data and parameters. | NA - No novel base-model training or fine-tuning documented |
| Prompt Engineering | 5a | Describe the evolution of study prompt development. | pp. 5, 7 (§§2.4, 2.5)<br>P - Prompt strategy/freeze, not evolution |
| Prompt Engineering | 5ai | Describe the sources of prompts. | pp. 5–7 (§§2.4–2.5); Supp. Table 2<br>P - Task templates and input sources, not prompt-design provenance |
| Prompt Engineering | 5aii | State the number and characteristics of the individual(s) involved in prompt engineering. | NR - Prompt engineers not identified |
| Prompt Engineering | 5aiii | Provide details of any patient and public involvement during prompt engineering. | NR - Public involvement in prompt design not reported |
| Prompt Engineering | 5b | Provide study prompts. | pp. 5–7 (§§2.4–2.5); Supp. Table 2<br>P - Representative prompts only, not all prompts |
| Query Strategy | 6a | State route of access to generative AI model. | p. 9 (§2.8)<br>R - API access documented |
| Query Strategy | 6b | State the date(s) and location(s) of queries for the generative AI-driven chatbot(s) including the day, month, and year as well as city and country. | NR - Dates and locations of API queries not reported |
| Query Strategy | 6c | Describe whether prompts were input into separate chat session(s). | NR - Separate-session design not documented |
| Query Strategy | 6d | Provide all generative AI-driven chatbot output/responses | pp. 9, 14 (§§2.8, 8)<br>P - Outputs mentioned in repository, full response set not established |
| Prompting | 7a | Define the ground truth or reference standard used to define successful generative AI-driven chatbot performance. | p. 5 (§2.3)<br>R - Senior-reviewed consensus expert reference standard |
| Prompting | 7b | Describe the process undertaken for generative AI-driven chatbot performance evaluation. | pp. 8–11 (§§2.6–2.7, 3.2–3.5)<br>R - Methods and agreement results |
| Prompting | 7bi | State the number and characteristics of team members involved in performance evaluation. | pp. 5–6 (§§2.3, 2.4.5)<br>P - Assessor roles, limited personnel characteristics |
| Prompting | 7bii | Provide details of any patients and public involvement during the evaluation process. | NR - Public involvement in evaluation not reported |
| Prompting | 7biii | State whether evaluators were blinded to the identity of the generative AI-driven chatbot(s) under assessment. | NR - Blinding of evaluators not reported |
| Sample Size | 8 | Report how the sample size was determined. | p. 4 (§2.2)<br>P - Selection by feasibility; formal sample size rationale not provided |
| Data Analysis | 9a | Describe statistical analysis methods, including any evaluation of reproducibility of generative AI-driven chatbot responses. | pp. 8–9 (§2.7)<br>P - Statistics described; response reproducibility not tested |
| Data Analysis | 9ai | Report the measures used for performance evaluation. | pp. 8–9 (§§2.6–2.7)<br>R - Agreement and error metrics |
| Results | Results | Results | |
| | 10a | Report the performance evaluation undertaken including the alignment between generative AI-driven chatbot output and ground truth or reference standard using quantitative or mixed methods approaches as applicable. | pp. 10–11 (§§3.2–3.5); Tables 1–4<br>R - Agreement to gold standard |
| | 10b | For responses deviating from the ground truth or reference standard, state the nature of the difference(s). | pp. 10–12 (§§3.3, 4.3); Table 2<br>R - Direction, distribution and category of errors |
| | 10c | Indicate where the study protocol can be accessed or state that a protocol was not prepared. | NR - Study protocol availability not stated |
| Discussion | Discussion | Discussion | |
| | 11a | Interpret study findings in the context of relevant evidence. | pp. 11–12 (§§4.1–4.3)<br>R - Interpretation in prior evidence context |
| | 11b | Describe the strengths and limitations of the study. | p. 13 (§4.4)<br>R - Strengths and limitations |
| | 11c | Describe the potential implications for practice, education, policy, regulation, and research. | p. 13 (§§4.5–4.6)<br>R - Implications and future work |
| Open Science | Open Science | Open Science | |
| Disclosures | 12a | Report any relevant conflicts of interest for all authors. | p. 14 (§5)<br>R - Conflicts statement |
| Funding | 12b | Report sources of funding and their role in the conduct and reporting of the study. | p. 14 (§7)<br>R - Funding statement |
| Ethics | 12c | Describe the process undertaken for ethical approval. | NR - No ethics approval/waiver statement |
| Ethics | 12ci | Describe the measures taken to safeguard data privacy of patient health information, as applicable. | NR - Data privacy safeguards not described |
| Ethics | 12cii | State whether permission/licensing was obtained for the use of original, copyrighted data. | NR - Licensing/permission statement absent |
| Protocol | 12d | Provide a study protocol. | NR - No study protocol linked or declared absent |
| Data availability | 12e | State where study data, code repository, and model parameters can be accessed. | pp. 9, 14 (§§2.8, 8)<br>P - Data/code repository cited; model parameter access not covered |

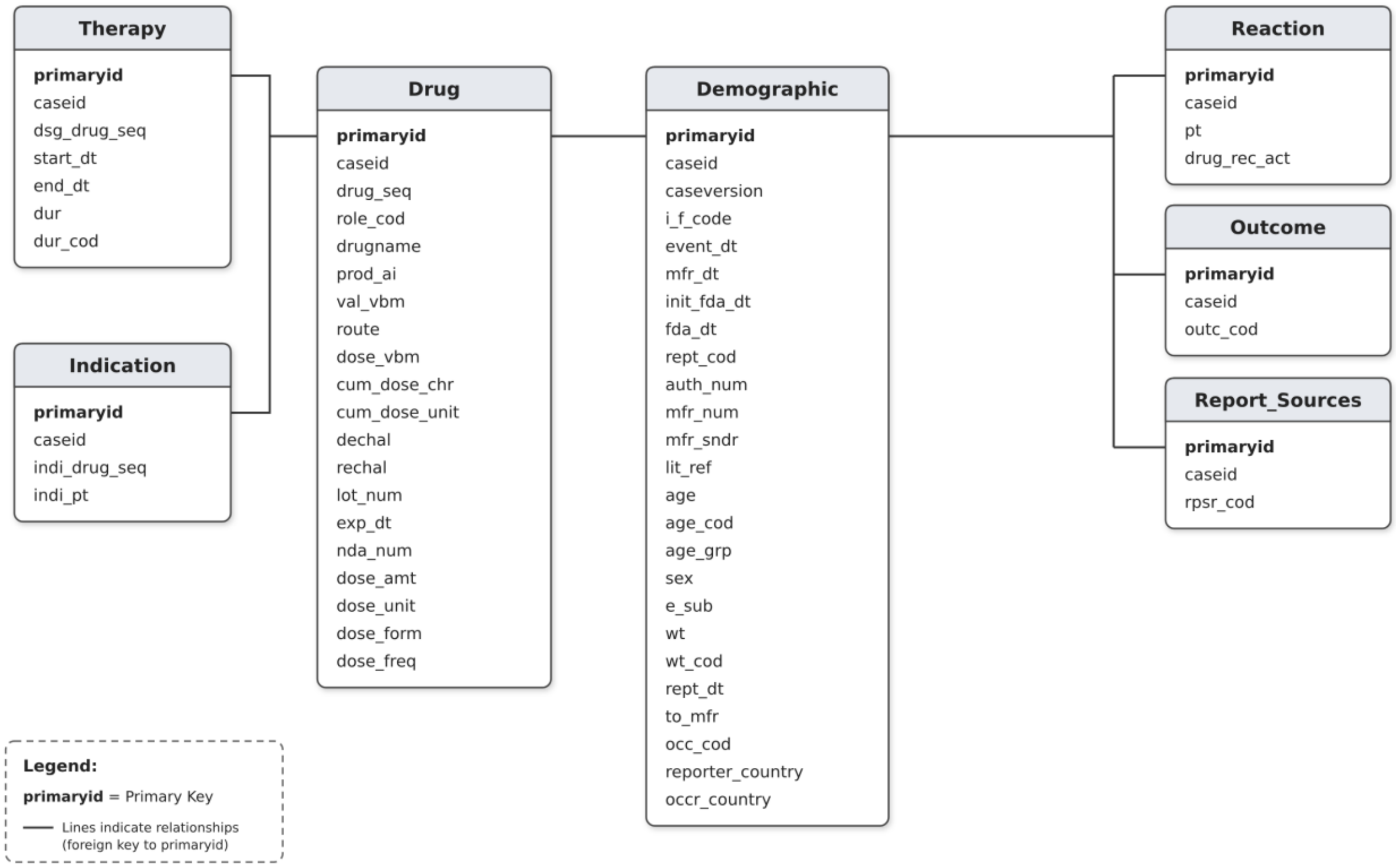

**Supplementary Figure 1. Relational Structure of the FAERS Database.** The FAERS database comprises interconnected tables capturing patient demographics, drug exposure, therapy, indications, reported reactions, clinical outcomes, and report sources. These tables are linked through shared identifiers to enable integration of case-level adverse event data.

*Legend: AE = adverse event; CASEID = case identification number; FAERS = FDA Adverse Event Reporting System; MedDRA = Medical Dictionary for Regulatory Activities; PT = Preferred Term; SEQN = sequence number; SOC = System Organ Class.*

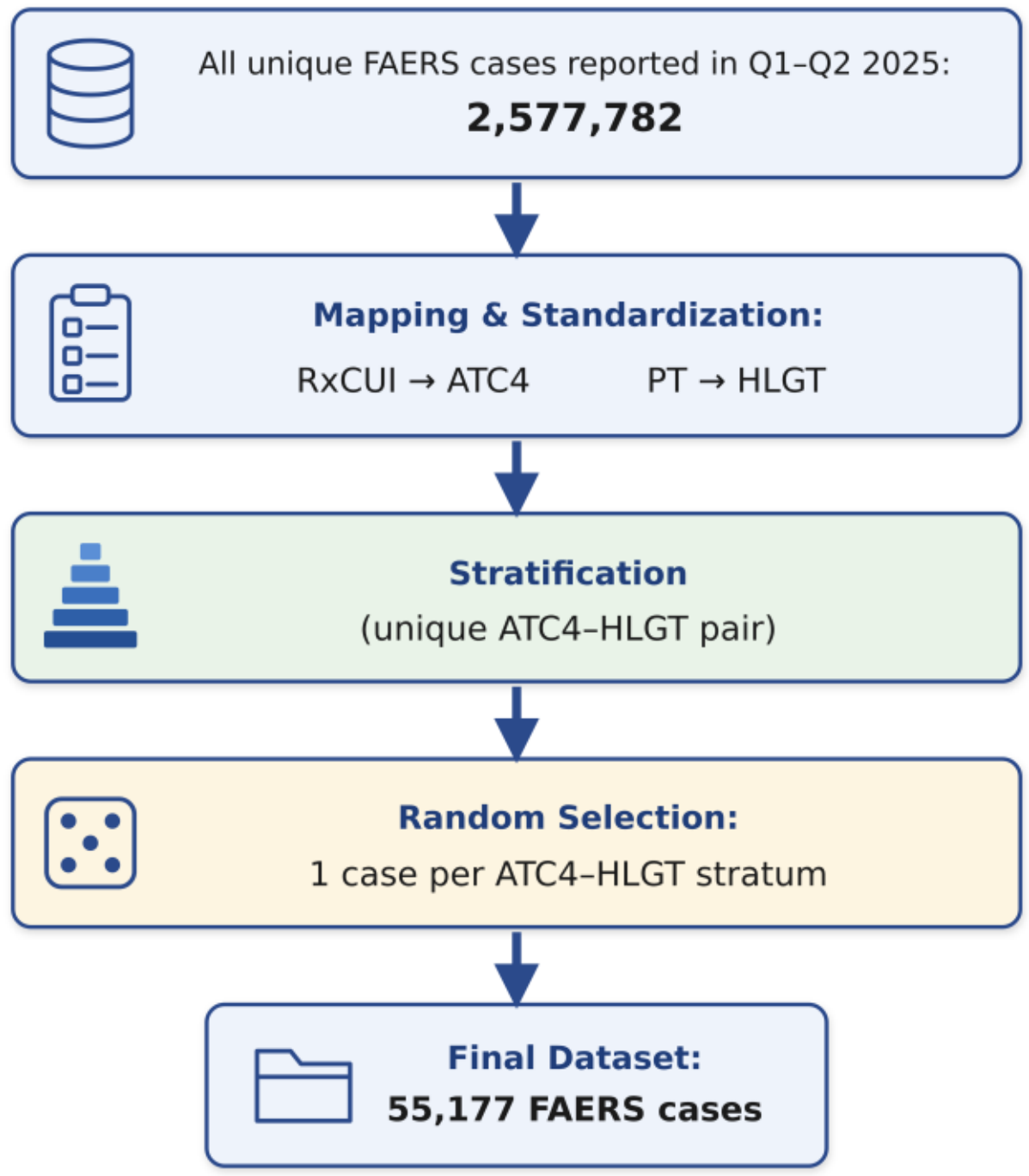

**Supplementary Figure 2. Workflow for Stratified Down-Sampling of FAERS Q1-Q2 2025 Data.** FAERS reports from Q1–Q2 2025 were standardized using drug and adverse event terminology mapping, stratified by unique ATC4–HLGT combinations, and down-sampled to one representative case per stratum to create the expert-assessed dataset.

*Legend: AE = adverse event; ATC4 = Anatomical Therapeutic Chemical Classification System, Level 4; FAERS = FDA Adverse Event Reporting System; HLGT = High Level Group Term; MedDRA = Medical Dictionary for Regulatory Activities; PT = Preferred Term; Q1–Q2 2025 = first and second quarters of 2025; RxCUI = RxNorm Concept Unique Identifier.*

*Disclaimer: The figure is AI-generated.*

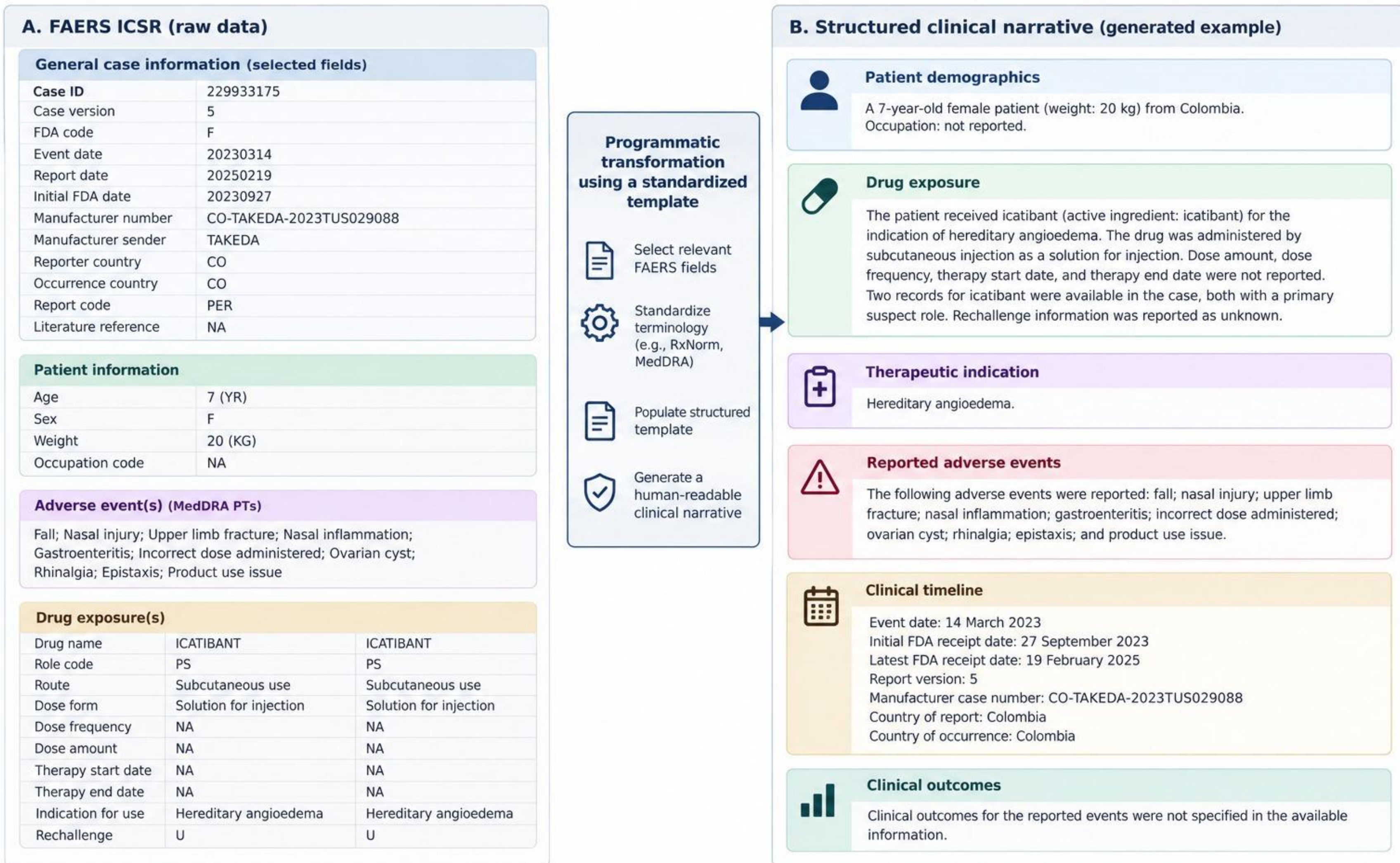

**Supplementary Figure 3. Retrieval of ICSRs and generation of standardized structured clinical narratives for human expert and MARCA-based causality assessment.** ICSR retrieval based on sampled ATC level 4–HLGT combinations and programmatic generation of standardized clinical narratives for human expert and MARCA assessment.

*Legend: ICSR = individual case safety report, MARCA = Multi-Agent Retrieval-Augmented Causality Assessment.*
*Disclaimer: The figure is AI-generated.*

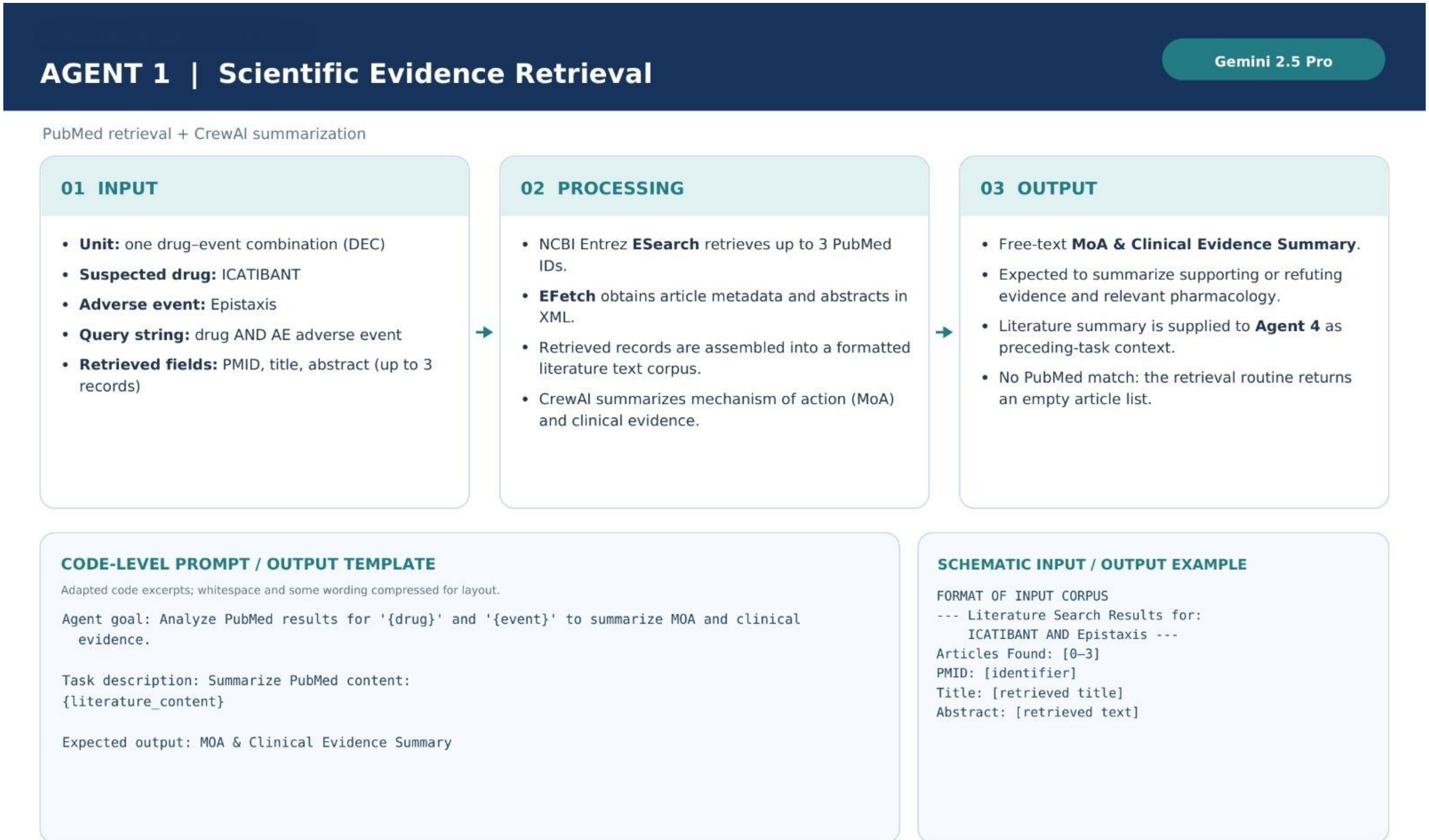

**Supplementary Figure 4. Agent 1 - Scientific Literature Retrieval and Synthesis.** PubMed evidence retrieval via the NCBI Entrez E-utilities API, standardized literature corpus generation, and LLM-based synthesis of drug mechanisms of action and clinical evidence to support Agent 4 causality assessment.

*<u>Disclaimer</u>: The figure is AI-generated.*

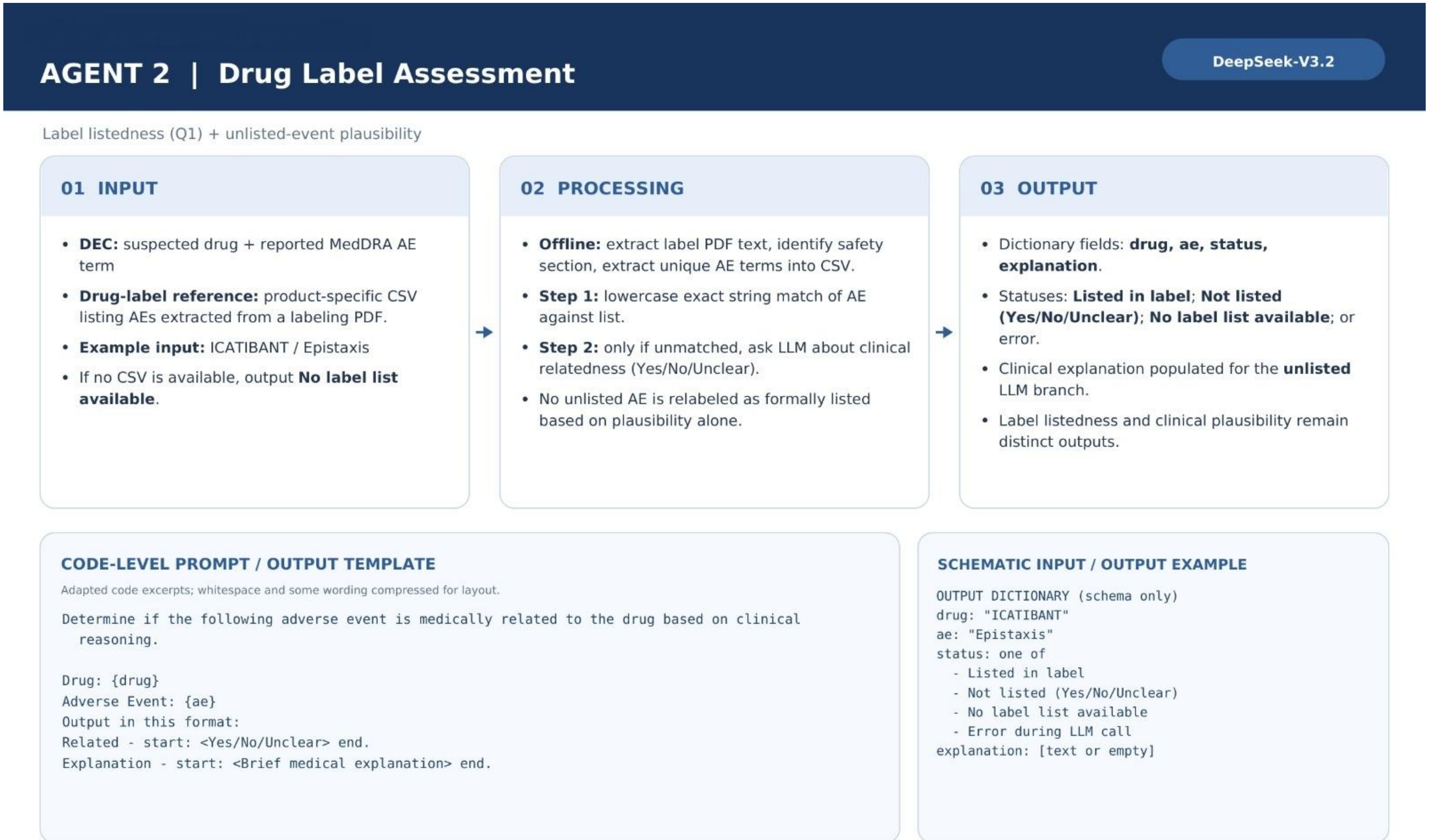

**Supplementary Figure 5. Agent 2 Drug Label Assessment.** Comparison of reported adverse events with drug-specific reference lists extracted from FDA-approved product labeling, followed by LLM-based clinical assessment of unmatched events and assignment of standardized output categories.
*Disclaimer: The figure is AI-generated.*

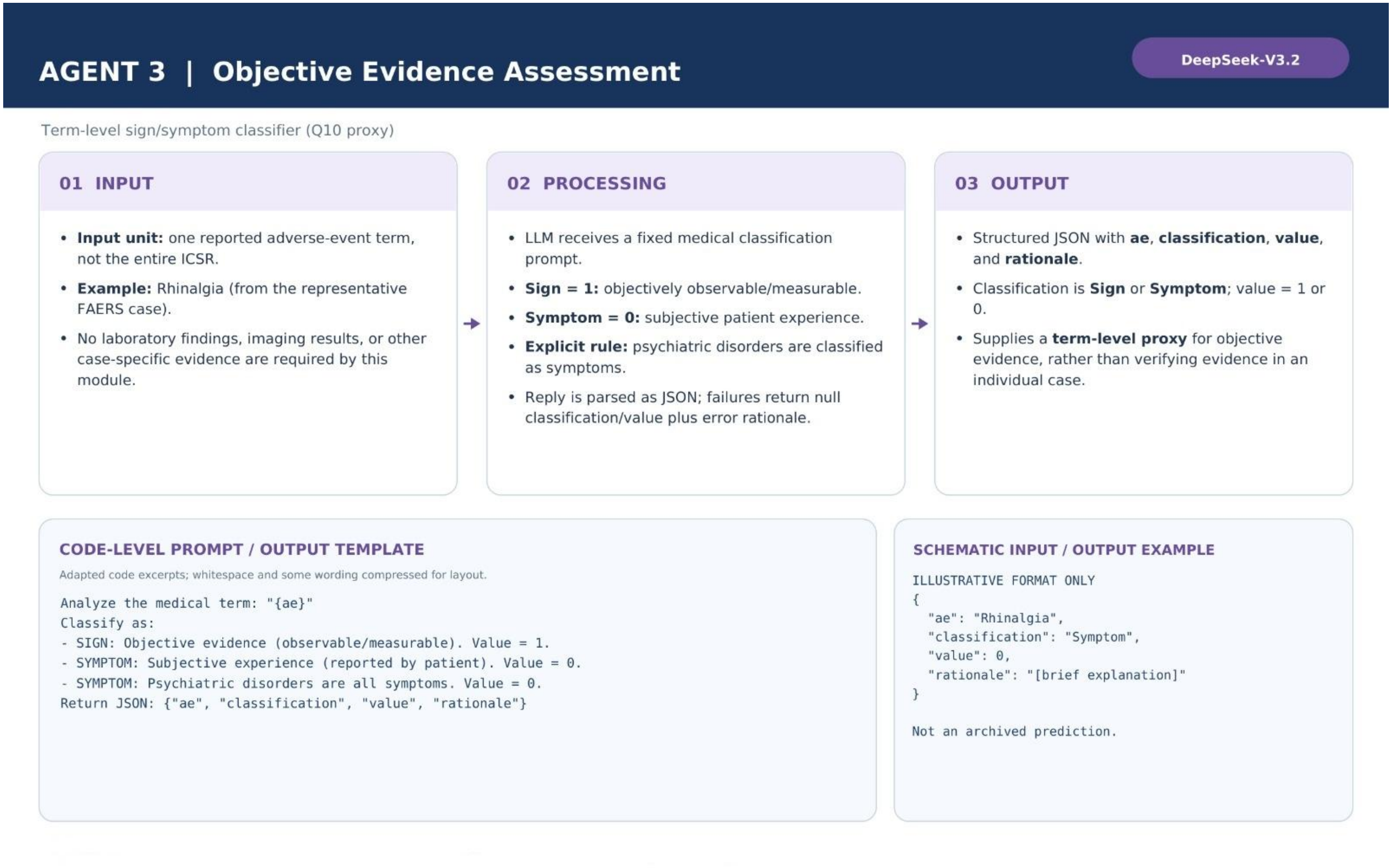

**Supplementary Figure 6. Agent 3 - Objective Evidence Assessment.** Structured LLM-based classification of reported adverse events as signs or symptoms, with corresponding numerical values and medical rationales to support Naranjo Question 10.

*<u>Disclaimer</u>: The figure is AI-generated.*

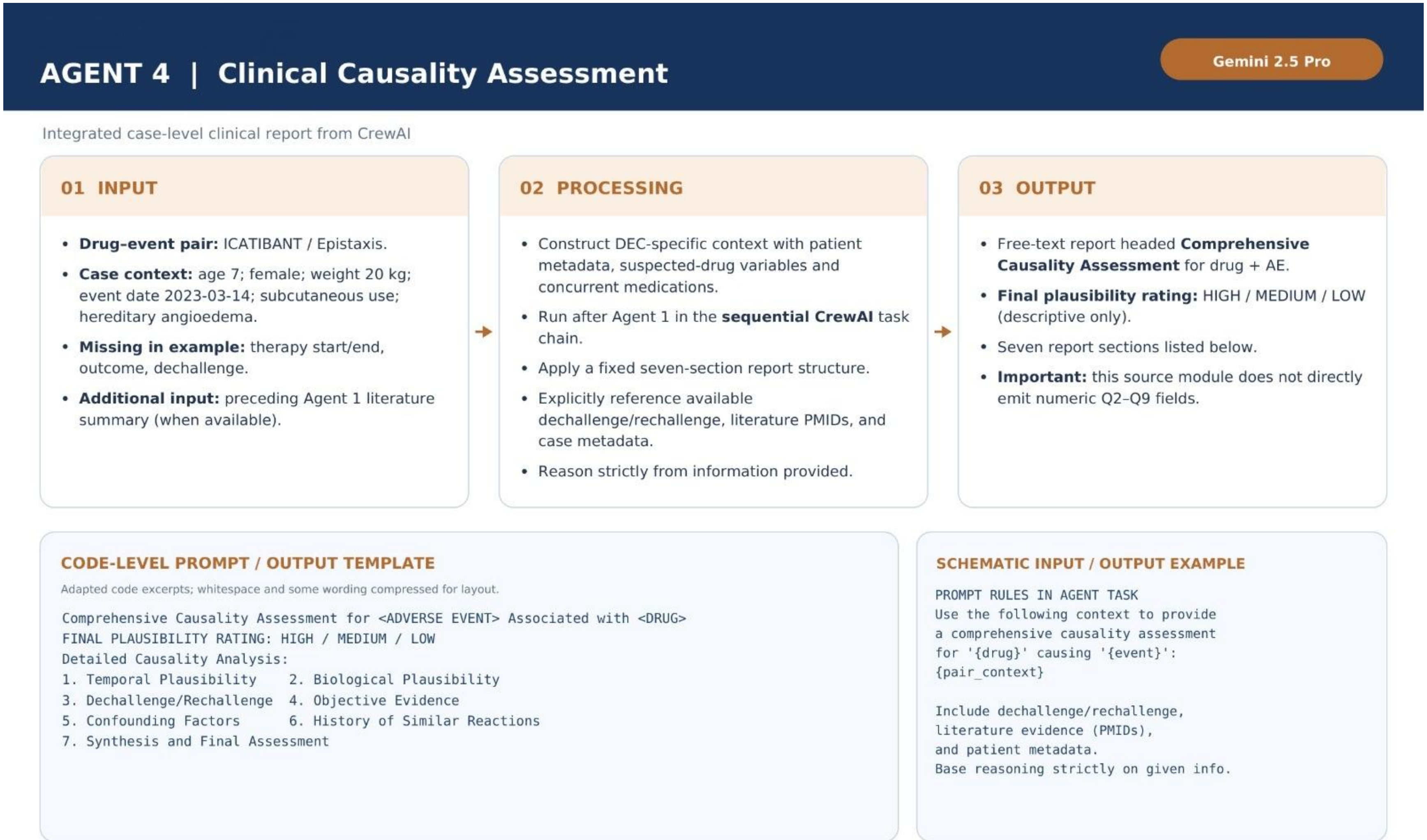

**Supplementary Figure 7. Agent 4 - Clinical Causality Assessment.** Integration of structured clinical case information and Agent 1's literature synthesis to generate standardized causality reports addressing temporal and biological plausibility, dechallenge and rechallenge, objective evidence, confounding factors, and overall clinical plausibility.

*<u>Disclaimer</u>: The figure is AI-generated.*

# Standard Operating Procedure

*Human expert causality assessment of FAERS drug-event combinations using the Naranjo algorithm*

| | |
|---|---|
| **Document ID** | **MARCA-PV-CA-001** |
| **Status / version** | Approved v0.1 |

## 1. Purpose and applicability

To standardize manual, question-level causality assessments and the creation of an adjudicated human reference standard for the MARCA research study. The unit of assessment is one specific suspected drug-adverse event combination (DEC), not the whole ICSR. The MARCA repository describes 215 FAERS ICSRs yielding 1,562 DECs in the human-reference set. This procedure applies to retrospective evaluation of those structured case narratives; no direct patient contact or intervention is permitted.

## 2. Personnel and responsibilities

- Junior assessor: reviews the structured narrative and permitted supporting sources, completes Q1-Q10 independently, records sources and clinical reasoning, and signs/dates the first-pass assessment.
- Senior pharmacovigilance expert: reviews every first-pass assessment and the available evidence, documents changes or unresolved ambiguities, and participates in case-by-case reconciliation.
- Study custodian (Senior pharmacovigilance expert): maintains version-controlled narratives, controls access to model predictions, checks completeness, freezes approved consensus records, and retains the audit trail.

## 3. Case materials and evidence standards

Assessors must use the same version of the structured FAERS narrative, including available demographics, suspected and concomitant medications, drug indications, dose and exposure dates, AE terms, onset/treatment timelines, clinical outcomes, and reported tests. The PRIMARYID identifies a specific FAERS report version; if multiple versions of an underlying case are present, the version-selection policy must be documented before review.

Any external FDA label or biomedical publication consulted for Q1 or mechanistic context must be logged with its bibliographic identifier/URL, version or publication date, and access date. External information may support general drug-event plausibility but cannot substitute for missing patient-specific dates, negative rechallenge evidence, laboratory findings, or alternative diagnoses. MARCA predictions must remain hidden until human consensus has been finalized.

## 4. Assessment workflow

1. Verify the ICSR identifier, version, suspected drug name, and MedDRA Preferred Term for the AE. Create a separate entry for each DEC, including multiple AEs for one drug or multiple suspected drugs for one AE.
2. Read the full narrative once before scoring; extract positive findings, negatives explicitly reported, chronology, concomitant treatments, comorbidities, and objective investigations.
3. For each of the 10 Naranjo questions, choose Yes, No, or Do not know; assign the scale-defined number; enter one brief clinical rationale and the supporting narrative excerpt or source reference.

4. Use Do not know when information is absent, contradictory, or insufficient. Do not treat an unreported action (e.g., no documented rechallenge or placebo administration) as an observed negative result.
5. Sum the 10 numerical scores and derive the final category using the official cutoffs. Record any uncertainty; do not modify scores to fit a desired final category.
6. Submit the first-pass record to the senior expert, reconcile each disputed response, and freeze the signed consensus assessment before comparing it with model predictions.

**5. Naranjo scoring matrix**

Score each item for the specific DEC. "Do not know" always scores 0 but must remain distinct from an observed "No" in the extraction form.

| Item | Naranjo question (condensed) | Yes | No | Unknown |
|---|---|---|---|---|
| **Q1** | Prior conclusive evidence for the drug–AE association | +1 | 0 | 0 |
| **Q2** | AE began after suspected drug administration | +2 | −1 | 0 |
| **Q3** | AE improved after withdrawal or a specific antagonist | +1 | 0 | 0 |
| **Q4** | AE recurred after readministration (rechallenge) | +2 | −1 | 0 |
| **Q5** | Other causes could independently explain the AE | −1 | +2 | 0 |
| **Q6** | AE recurred following placebo administration | −1 | +1 | 0 |
| **Q7** | Toxic drug concentration documented in blood/fluids | +1 | 0 | 0 |
| **Q8** | AE severity varied with drug dose | +1 | 0 | 0 |
| **Q9** | Similar AE on previous exposure to same/similar drug | +1 | 0 | 0 |
| **Q10** | AE confirmed by objective evidence | +1 | 0 | 0 |

**6. Item-specific decision rules**

**Q1 - Prior reports.** Use a documented source indicating a recognized drug–AE association, preferably regulatory product information or credible scientific evidence. MARCA uses FDA-label assessment for Q1; specify explicitly whether label listedness is accepted as the study's operational proxy for "previous conclusive reports." Not finding an AE in a searched label is not, by itself, proof that no conclusive reports exist.

**Q2 - Temporal sequence.** Assign Yes (+2) only if the available timeline supports AE onset after exposure; consider plausible latency when interpreting the sequence. Assign No (−1) only if the AE clearly predates first relevant exposure. If drug start/onset timing is unknown, contradictory, or too imprecise to establish order, assign Do not know (0); never infer dates from general pharmacology alone.

**Q3 - Dechallenge.** Assign Yes (+1) if improvement after documented drug discontinuation or specific antagonist is reported. If the outcome after withdrawal is explicitly not improved, use No (0). If withdrawal or post-withdrawal course is absent or ambiguous, use Do not know (0).

**Q4 - Rechallenge.** Use Yes (+2) only with documented re-exposure and recurrence; No (−1) requires documented re-exposure without recurrence. If no rechallenge is recorded, use Do not know (0). No rechallenge should be undertaken for research purposes.

**Q5 - Alternative causes.** Review concomitant medications, underlying diseases, presenting illness, infections, procedures, and other clinically plausible etiologies. Yes (−1): at least one documented alternative could independently cause the AE. No (+2): available case-specific evidence adequately supports the absence or exclusion of credible alternatives. Do not know (0): information is insufficient to determine whether alternatives exist. Absence of reported comorbidities or concomitant drugs is NOT sufficient by itself to assign No (+2). Log the competing explanation and supporting facts.

**Q6 - Placebo.** Yes (−1) or No (+1) requires a reported placebo challenge and its observed result. Where there is no placebo exposure, use Do not know (0).

**Q7 - Toxic concentration.** Yes (+1) requires a documented drug concentration in a fluid together with a known toxic range. If testing confirms nontoxic concentrations, use No (0); if no relevant measurement is reported, use Do not know (0).

**Q8 - Dose response.** Yes (+1) requires documented dose escalation/worsening or reduction/improvement with relevant chronology. If tested dose changes showed no relationship, use No (0); otherwise Do not know (0).

**Q9 - Prior similar reaction.** Yes (+1) requires a documented similar reaction to the same or pharmacologically related drug in a previous exposure. If explicit prior exposure without such reaction is reported, use No (0); otherwise Do not know (0).

**Q10 - Objective confirmation.** Yes (+1) requires documented corroboration of the AE by an examination finding, laboratory result, imaging study, pathology, instrument-measured parameter, or similar verifiable observation. A reported MedDRA PT, diagnostic label, or subjective symptom alone does not establish objective confirmation. Use No (0) for explicitly absent objective corroboration after evaluation and Do not know (0) if no objective investigation or finding is described. Name the evidence in the rationale.

### 7. Overall classification

Calculate the total as the arithmetic sum of Q1–Q10. Classify each DEC as definite (≥9), probable (5–8), possible (1–4), or doubtful (≤0). Preserve question-level responses, scores, rationale, and total; the same ICSR may yield different classes for different DECs.

### 8. Senior review and reconciliation

- The senior expert verifies the case/drug/event pairing, evidence provenance, response category, score mapping, and arithmetic for every DEC.
- Any proposed correction is logged with question number, junior response/score, senior proposed response/score, specific supporting evidence, and final resolution. The junior and senior discuss discrepant items and document the agreed final score and rationale.
- If disagreements cannot be resolved, refer the DEC to a predefined escalation process or mark it unresolved; do not silently impose consensus. The actual escalation pathway and whether unresolved DECs were included must be confirmed from study records.
- Retain initial assessments separately from final consensus and record reviewer identity/date. Report initial inter-rater agreement only if genuinely independent dual assessments were collected; post-reconciliation consensus is not the same metric.

### 9. Data quality, independence, and audit trail

- Run validation checks for all 10 responses and allowable numerical scores, nonmissing rationales, correctly summed totals, and class thresholds; validate uniqueness at the DEC identifier level.
- Lock the input narrative version and document source cutoffs and label retrieval dates. No new information may be added to the human assessment from MARCA outputs.
- Keep the evaluation gold standard separate from any expert-annotated knowledge base exposed to MARCA during evaluation. Identify and document any overlapping cases, labels, or records before reporting an unbiased held-out comparison.
- Freeze final consensus records and preserve the audit log and source identifiers according to the study's approved data-governance plan.

### 10. Structured assessment form (per DEC)

A standardized assessment dataset was used to document human expert causality assessments and compare them with MARCA-generated outputs. Each row represented an individual DEC, uniquely identified by a record identifier (id) and linked to the corresponding FAERS case (case_id).

#### A.1. Case identification and clinical information

| Field | Description |
|---|---|
| id | Unique assessment record identifier |
| case_id | Identifier linking the DEC to the corresponding FAERS case |
| drug_name | Suspected drug associated with the assessed event |
| pt | Adverse event coded as a MedDRA Preferred Term |
| case_narrative | Structured clinical narrative containing the available patient, drug exposure, adverse event, and clinical timeline information (see supplementary figure 3) |
| dec | Supplementary DEC-related metadata recorded in the dataset |

#### A.2. Question-level causality assessment

For each DEC, all ten Naranjo questions were evaluated individually. The human expert assessment consisted of a numerical score and a supporting clinical rationale for each question. MARCA-generated scores and rationales were stored in corresponding fields, together with question-level agreement indicators.

| Naranjo question | Human expert score | Human clinical rationale | MARCA score | MARCA rationale | Agreement |
|---|---|---|---|---|---|
| Q1 | q1_score_human | q1_reasoning_human | q1_score_ai | q1_reasoning_ai | q1_agreement |
| Q2 | q2_score_human | q2_reasoning_human | q2_score_ai | q2_reasoning_ai | q2_agreement |
| Q3 | q3_score_human | q3_reasoning_human | q3_score_ai | q3_reasoning_ai | q3_agreement |
| Q4 | q4_score_human | q4_reasoning_human | q4_score_ai | q4_reasoning_ai | q4_agreement |
| Q5 | q5_score_human | q5_reasoning_human | q5_score_ai | q5_reasoning_ai | q5_agreement |
| Q6 | q6_score_human | q6_reasoning_human | q6_score_ai | q6_reasoning_ai | q6_agreement |
| Q7 | q7_score_human | q7_reasoning_human | q7_score_ai | q7_reasoning_ai | q7_agreement |
| Q8 | q8_score_human | q8_reasoning_human | q8_score_ai | q8_reasoning_ai | q8_agreement |
| Q9 | q9_score_human | q9_reasoning_human | q9_score_ai | q9_reasoning_ai | q9_agreement |
| Q10 | q10_score_human | q10_reasoning_human | q10_score_ai | q10_reasoning_ai | q10_agreement |

Each human expert rationale documented the evidence supporting the assigned score, including relevant information from the clinical narrative or external reference sources when applicable.

### A.3. Final causality scores and agreement

| Field | Description |
|---|---|
| score_human | Total Naranjo score assigned by the human expert assessment |
| score_ai | Total Naranjo score generated by MARCA |
| score_agreement | Agreement indicator between the final human and MARCA numerical scores |

Final causality categories were determined from the total Naranjo scores according to the standard classification thresholds: Definite (≥9), Probable (5–8), Possible (1–4), and Doubtful (≤0).

### A.4. Supplementary MARCA-generated evidence

The dataset also retained additional MARCA-generated outputs to support evaluation of the framework:

| Field | Description |
|---|---|
| q1_ai_answer | Drug label assessment output supporting Q1 |
| literature_ai_answer | Scientific literature retrieval and summarization output |
| moa_ai_answer | Mechanism-of-action and pharmacological evidence summary |
| questions2_9_ai_answer | Clinical causality assessment output covering Q2–Q9 |

These fields provided supplementary information on the evidence retrieved and reasoning generated by the framework.

### A.5. Human expert assessment and reconciliation

The human expert assessments were conducted before comparison with MARCA-generated results. Following the initial assessment, the assigned Naranjo scores and supporting rationales were reviewed by a senior pharmacovigilance expert and reconciled to establish the final reference standard.

The reconciled human expert scores and clinical rationales were represented by the q1_score_human–q10_score_human and q1_reasoning_human–q10_reasoning_human fields, respectively.

MARCA-generated results were subsequently compared with the finalized human expert assessments at both the individual-question and total-score levels. Agreement was recorded using the corresponding question-level and overall agreement fields.

## 11. Pre-finalization checks

☐ One completed assessment per intended DEC, correct pairings and narrative version.

☐ Complete Q1–Q10 responses, permitted scores, and concise source-based rationales.

☐ No unreported rechallenge/placebo or unmeasured drug level incorrectly scored as an observed No.

☐ Chronology, alternative etiologies, and objective evidence explicitly considered.

☐ Arithmetic, thresholds, and classification verified.

☐ All senior corrections and final reconciliations traceable.

☐ Reference standard frozen prior to model evaluation, and overlap with the RAG knowledge base audited.

## References and provenance